\documentclass{article}

\usepackage{PRIMEarxiv}

\usepackage[utf8]{inputenc} 
\usepackage[T1]{fontenc}    
\usepackage{hyperref}       
\usepackage{url}            
\usepackage{booktabs}       
\usepackage{amsfonts}       
\usepackage{nicefrac}       
\usepackage{microtype}      
\usepackage{lipsum}
\usepackage{graphicx}
\usepackage[numbers,sort&compress]{natbib}          
\usepackage{amssymb}
\usepackage{amsmath}
\usepackage{tabto}
\usepackage[english]{babel}
\usepackage{blindtext}
\usepackage[T1]{fontenc}
\usepackage{amsthm}
\usepackage{multirow}
\usepackage{tabu}
\usepackage{booktabs}
\usepackage[utf8]{inputenc} 
\usepackage{amsfonts}       
\usepackage{graphicx}
\usepackage{bm}
\usepackage[switch]{lineno}
\usepackage{comment}
\usepackage{float} 
\usepackage{algorithmic}
\usepackage{subcaption} 
\usepackage{textcomp}
\usepackage{soul, color}
\usepackage{url}
\usepackage{multirow}
\usepackage[ruled,vlined]{algorithm2e} 
\usepackage[normalem]{ulem}
\usepackage{xcolor}
\usepackage{pifont}   
\usepackage{tikz}
\usepackage{threeparttable}
\usepackage{rotating}
\usepackage{afterpage}
\usepackage{placeins}

\graphicspath{{media/}}     

\definecolor{myYellow}{HTML}{E8D427}
\newcommand{\ybullet}{%
\tikz[baseline=-0.6ex] \draw[draw=black, line width=0.6pt, fill=myYellow] (0,0) circle (0.95ex);%
}
\newcommand{\bluecirc}{%
\tikz[baseline=-0.6ex] \draw[line width=0.9pt, draw=blue!90!black] (0,0) circle (0.9ex);%
}

\newcommand{\bluecircgreendot}{%
\tikz[baseline=-0.6ex]{
  \draw[line width=0.9pt, draw=blue!90!black] (0,0) circle (0.9ex);
  \fill[green!70!black] (0,0) circle (0.45ex);
}%
}
  
\title{Weakly Supervised Patch Annotation for Improved Screening of Diabetic Retinopathy
}

\author{
  Shramana Dey \\
  Indian Statistical Institute \\
  Kolkata, West Bengal, India \\
  \texttt{shramanadey\_r@isical.ac.in, shramanadey96@gmail.com}
  \And
  Abhirup Banerjee \\
  Department of Engineering Science, University of Oxford \\
  Radcliffe Department of Medicine, University of Oxford \\
  Oxford, United Kingdom \\
  \texttt{abhirup.banerjee@eng.ox.ac.uk}
  \And
  B. Uma Shankar \\
  Techno India University \\
  Kolkata, West Bengal, India \\
  \texttt{uma@isical.ac.in}
  \And
  Ramachandran Rajalakshmi \\
  Madras Diabetes Research Foundation \\
  Chennai, Tamil Nadu, India \\
  \texttt{drraj@drmohans.com}
  \And
  Sushmita Mitra \\
  Indian Statistical Institute \\
  Kolkata, West Bengal, India \\
  \texttt{sushmita@isical.ac.in}
}

\begin{document}
\maketitle

\begin{abstract}
Diabetic Retinopathy (DR) requires timely screening to prevent irreversible vision loss. However, its early detection remains a significant challenge since often the subtle pathological manifestations (lesions) get overlooked due to insufficient annotation. Existing literature primarily focuses on image-level supervision, weakly-supervised localization, and clustering-based representation learning, which fail to systematically annotate unlabeled lesion region(s) for refining the dataset. Expert-driven lesion annotation is labor-intensive and often incomplete, limiting the performance of deep learning models. We introduce \textbf{S}imilarity-based \textbf{A}nnotation via \textbf{F}eature-space \textbf{E}nsemble (SAFE), a two-stage framework that unifies weak supervision, contrastive learning, and patch-wise embedding inference, to systematically expand sparse annotations in the pathology. SAFE preserves fine-grained details of the lesion(s) under partial clinical supervision. In the first stage, a dual-arm Patch Embedding Network learns semantically structured, class-discriminative embeddings from expert annotated patches. Next, an ensemble of independent embedding spaces extrapolates labels to the unannotated regions based on spatial and semantic proximity. An abstention mechanism ensures trade-off between highly reliable annotation and noisy coverage. Experimental results demonstrate reliable separation of healthy and diseased patches, achieving upto 0.9886 accuracy. The annotation generated from SAFE substantially improves downstream tasks such as DR classification, demonstrating a substantial increase in F1-score of the diseased class and a performance gain as high as 0.545 in Area Under the Precision-Recall Curve (AUPRC). Qualitative analysis, with explainability, confirms that SAFE focuses on clinically relevant lesion patterns; and is further validated by ophthalmologists.
\end{abstract}


\section{Introduction}
\label{sec:intro}

\begin{figure*}[h]
    \centering
    \includegraphics[width=0.6\textwidth]{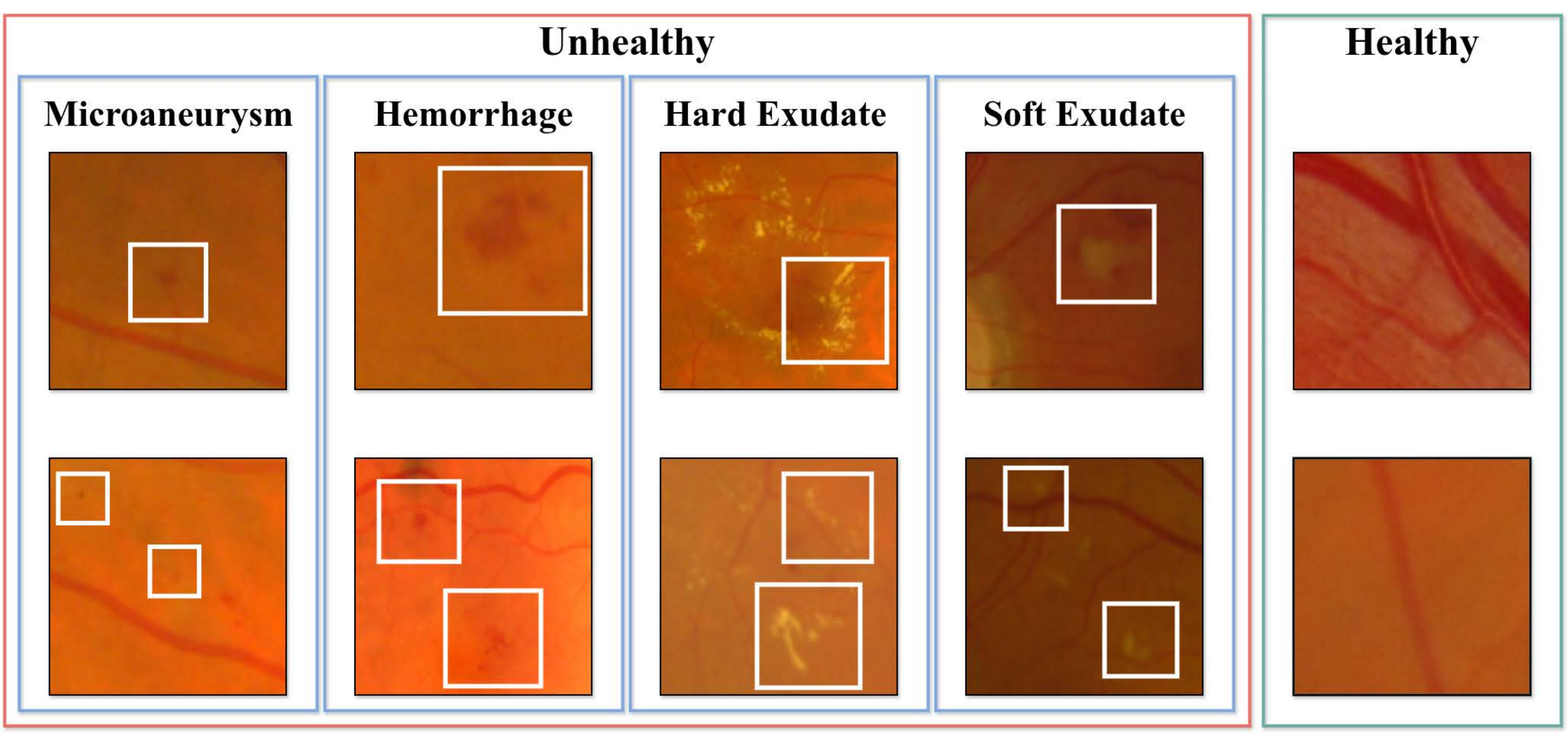}
    \caption{Representative patches from fundus images of Messidor dataset, illustrating key DR-related pathologies and healthy retinal regions. White bounding boxes highlight the lesions associated with DR.}
    \label{fig:lesions}
\end{figure*}

 Diabetic Retinopathy (DR) remains a leading cause of preventable vision loss \citep{raman2022prevalence, akhtar2025deep} worldwide. It progresses gradually through the non-proliferative stages--mild, moderate, severe--culminating in proliferative DR, which is characterized by neovascularization and increased risk of vision-threatening complications \citep{zhou2021benchmark}. Early stage manifestations of DR, such as microaneurysms and intraretinal hemorrhages, collectively referred to as red lesions \citep{lama2016redlesion}, are subtle and difficult to detect due to their low contrast and visual similarity to the surrounding retinal background. Bright lesions, including hard and soft exudates, are generally more distinct; the former appear as sharply defined yellow-white deposits, while the latter are more diffuse with poorly defined boundaries. Fig.~\ref{fig:lesions} illustrates the lesions of the DR spectrum.

As DR is often asymptomatic in its early stages, a timely and accurate diagnosis is essential to prevent irreversible retinal damage \citep{li2024current}. This has driven increasing interest in automated and scalable screening systems to reduce the reliance on overburdened clinical workflows \citep{rajesh2023artificial}. Recent advances in machine learning (ML) and deep learning (DL) have led to the development of effective DR screening models \citep{sun2021lesion, dey2023multi, zhang2024semantic}. Unlike ophthalmologists, who reason holistically using clinical context, the performance of the learning models is dependent on the availability of high-quality fine-grained annotations. This constrains the accuracy of decision-making by these models, to the extent and quality of labeled supervision \citep{hou2024clinical}. Most DR datasets suffer from non-uniform lesion distribution, resulting in partial annotation. Since the density of lesions  varies with the severity of the disease and detailed labeling is time consuming and labor intensive, many pathological regions remain unlabeled. Small lesions such as microaneurysms and dot-blot hemorrhages often go undetected, due to their subtle appearance and loss of information during associated downsampling. These limit the ability of supervised models to learn discriminative characteristics specific to the lesions. Conventional semi-supervised learning (SSL) \citep{he2024open, kim2024semi, zhao2025metassl} leverages unlabeled data, along with labeled data, to improve classification performance. However, its applicability remains limited in DR screening due to coarse lesion boundaries inadvertently including surrounding healthy regions to induce noisy labels.

Lesion segmentation \citep{kim2024semi, playout2019novel} relies on precise lesion annotation, which is available only for a few datasets such as IDRiD \citep{idrid} and e-ophtha \citep{decenciere2013teleophta}. Obtaining precise expert-annotated lesion segmentation at higher scales remains impractical, and is limited by sample size as well as lesion density and diversity. Segmentation models inherit bias from partial lesion coverage \citep{kim2024semi, yang2024anomaly} -- leading to incomplete feature learning by the model. Weakly supervised DR grading approaches, including attention-based frameworks \citep{zhou2019collaborative}, are often optimized for improving global classification accuracy. However, these paradigms operate on existing annotations, involving missing supervisory cues for enhanced performance. As a result, the gain in performance 
remains constrained in partial supervision. Explicit annotation expansion addresses such structural constraints to enable systematic refinement of the dataset. Recent literature has highlighted the value of lesion-level information in the grading of DR \citep{cao2022collaborative}. 

Researchers have explored diverse learning strategies to operate under weak, partial or missing supervision, in existing datasets. These collectively emphasize the value of structured representation learning under limited supervision. Although labeling of unlabeled data was found to be effective in natural images \citep{gudapati2025incremental}, the concept has been minimally explored in the medical domain. Weak supervision was employed on textual reports to annotate microaneurysms \citep{dai2017retinal},  highlighting the importance of indirect supervision to compensate for missing lesion annotation. However, it failed to generalize well across heterogeneous lesion categories. Prototypical networks \citep{snell2017prototypical}  demonstrated label-efficient learning via prototype matching, while Deep Cluster \citep{caron2018deep} learned domain-specific latent features without direct supervision. A dynamic Prototype-aware weakly supervised segmentation framework \citep{he2024open} represented classes using global centroid assumption in homogeneous distributions. As DR lesions exhibit heterogeneous morphology, it necessitates a more flexible approach for feature representation that does not rely on simple prototype learning.

Contrastive Learning frameworks, such as SimCLR \citep{chen2020simple}, demonstrated that representation alignment through similarity maximization yielded highly structured embedding spaces. Extending this, Supervised Contrastive Learning (SCL) \citep{khosla2020supervised} has been used in DR classification frameworks \citep{sun2021lesion, cheng2021prior, huang2021lesion} to improve discriminative capacity in the presence of sparse and imbalanced labels. Recent literature on contrastive learning \citep{hou2024clinical} explored disease diagnosis given low quality images. Enforcing intra-class compactness and inter-class separation in the embedding space, SCL facilitated robust feature learning under class imbalance to accommodate the high morphological variance of DR lesions \citep{juyal2024sc}. SCL allows a model to capture nuanced similarities between diverse lesion types. 

Efforts to address annotation limitations include an iterative label correction pipeline \citep{lin2025efficiency} and expert-in-the-loop querying to detect mislabeled regions \citep{bernhardt2022active}. However, they risk over-cleaning or involve high human involvement. Only a limited number of studies explicitly focused on refining annotation quality as a central objective. Existing literature acknowledges challenges posed by noisy or partial supervision and addresses them within task-specific models. Though they demonstrate promising performance gain, the benefits remain confined to the associated model architecture. A principled automated dataset refinement strategy is required to enable reusable and task agnostic enhancement of data quality.

We propose an automated framework -- \textbf{S}imilarity-based \textbf{A}nnotation via \textbf{F}eature-space \textbf{E}nsemble (SAFE) -- that infers reliable patch-level annotations under weak supervision. SAFE derives lesion-specific insights from disease-level labels and extrapolates partial annotations to produce completely annotated fundus images. The patch-based strategy preserves the resolution of the lesion(s) and allows the extraction of rich pathological representations. This enhances the quality of the DR dataset with minimal clinical effort. The resulting annotated data can be used to support various downstream tasks, like automated diagnosis or disease grading.

\begin{figure*}[ht]
    \centering
    \includegraphics[width=1.0\textwidth]{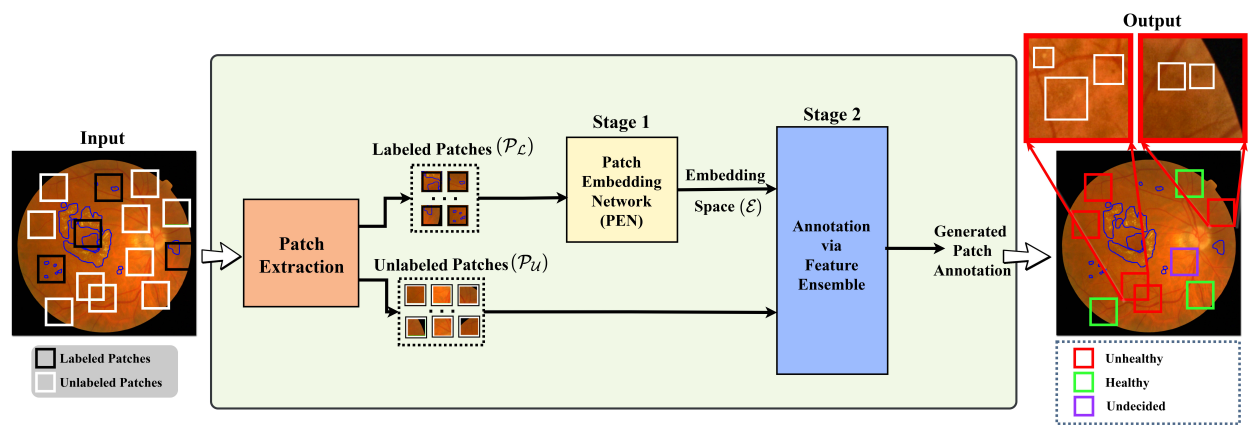}
    \caption{Overview of the SAFE framework for patch-wise label inference using weak supervision. Stage 1 takes labeled patches from weakly annotated fundus images to train multiple embedding models. Stage 2 uses the embedding spaces from Stage 1, together with unlabeled patches, to generate refined annotations for downstream tasks.}
    \label{fig:SAFESchematic}
\end{figure*}

A synergistic coupling of SSL with SCL is introduced into the SAFE framework, to maximize the utility of limited supervision. While SSL has become a standard way of modeling unlabeled data, it fails to handle noisy sample labels and class imbalance. To overcome these fundamental limitations, we integrate SCL to enforce discriminative and pathology-aware feature representations onto the latent space. A contrastive objective is applied to l2-normalized patch embeddings projected onto the unit hyper-sphere, with cosine similarity  enforcing the semantic structure in the learned space. This paradigm transforms the embedding space to a reliable foundation for automated lesion-centric annotation.
Contributions of this study are summarized below. 
\begin{itemize}
    \item A novel \textbf{S}imilarity-based \textbf{A}nnotation via \textbf{F}eature-space \textbf{E}nsemble (SAFE) framework  to unify weak supervision, contrastive learning, and patch-wise embedding inference for systematically expanding sparse lesion-level annotations. 
    \begin{itemize}
    \item The unique two-stage SAFE framework leverages a dual-arm Patch Embedding Network (PEN), in the first stage, \textbf{to learn discriminative embedding space from clinically annotated patches} -- into which unannotated patches are later projected for inference.
    \item An ensemble of multiple independent embedding spaces, in the second stage of SAFE, \textbf{extrapolates labels to unannotated regions} based on spatial and semantic proximity. This reduces model-specific biases, thus ensuring robust annotation. SAFE also allows \textbf{an abstention mechanism to ensure high-fidelity annotation} over noisy coverage.
    \item The new measures Decided Rate ($\text{D}_\text{rate}$) and extended Misclassification Rate ($MR$) explicitly quantify \textbf{annotation coverage and noise under abstention}. They \textbf{complement conventional metrics} by characterizing the extent and reliability of label propagation in weakly supervised settings.
       \end{itemize}
    \item Unlike most prior studies, SAFE operates at patch-level to \textbf{preserve the resolution of subtle minute lesions} that standard image-level downsampling typically loses. This enables the extraction of a rich pathological representation under weak clinical supervision. The design enables greater refinement of granular annotation,  with minimal manual input, making it suited for datasets having partial lesion coverage. 
     \item Explainability analysis \textbf{confirms that the embedding network focuses on pathological patterns} rather than background noise. This showcases the alignment of the inferred label(s) with clinical reasoning, and is further \textbf{validated by ophthalmic experts}.
    \item Incorporating \textbf{SAFE-inferred labels delivers substantial gains in DR datasets} for downstream tasks, including a performance gain as high as 0.545 in the Area Under the Precision-Recall Curve (AUPRC) on specific datasets. This establishes the effectiveness of downstream utility of our proposed SAFE framework.
\end{itemize}

The remainder of this manuscript is organized as follows. Section~\ref{sec:method} introduces the proposed SAFE framework, including problem formulation, architectural details, loss functions, datasets, and experimental protocol. Section \ref{sec:experimentaleval} presents comprehensive evaluations, both quantitative and qualitative, including an analysis of SAFE-inferred annotations in a downstream classification task. Finally, Section \ref{sec:conclusion} summarizes the key findings and concludes the article.

\begin{figure*}[ht]
    \centering
    \includegraphics[width=0.8\textwidth, trim=100 0 0 0, clip]{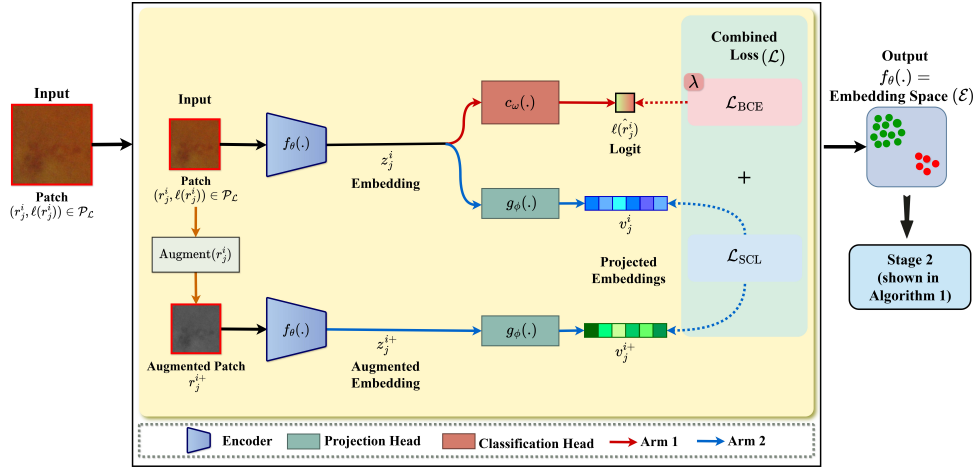}
    \caption{Overview of the patch embedding network (PEN) architecture. The labeled input patch and its corresponding augmented version are passed through the encoder $f_\theta(\cdot)$, and a combined loss is optimized. The resulting embedding space $(\mathcal{E})$ is utilized in the second stage of SAFE.}
    \label{fig:stage1Schematic}
\end{figure*}

\section{Methodology}
\label{sec:method}

This section outlines the problem, a description of patch extraction, and preliminary labeling strategies, followed by the SAFE objective and its two-stage annotation framework. It also enumerates the datasets used in this study, experimental details, and evaluation metrics.

\subsection{Problem formulation and initialization}
\label{sec: problemformulation}

The dataset $\mathcal{D} = \{(X_i, y_i, A_i)|~i=1,\cdots,N\}$ comprises $N$ fundus images, where each image $X_i \in \mathbb{R}^{H \times W \times C}$ carries a class label $y_i \in \{\text{No-DR},\allowbreak\ \text{DR}\}$. Here, $H$, $W$, and $C$ denote the height, width, and number of color channels, respectively, for each image. The class $DR$ includes four severity levels of DR, {\it viz} \text{Mild}, \text{Moderate}, \text{Severe}, \text{Proliferative}. $A_i$ denotes an annotation set, consisting of one or more annotation masks that mark distinct regions of interest (ROIs) indicating unhealthy areas. Formally,  
\begin{equation}
    A_i = \{ a^i_k | k=1,\cdots,K_i; a^i_k \subset X_i\},
\end{equation}
where $K_i$ is the number of unhealthy regions annotated in the image $X_i$. Each $a^i_k$ approximately bounds the region of occurrence of the lesion. All images $X_i$ $\in$ $\mathcal{D}$ labeled as \texttt{No-DR} are assigned an empty annotation set \(A_i ( = \emptyset)\), indicating the absence of observable abnormalities.

The objective of the proposed SAFE framework is to annotate an image $X_i$ at the patch level, using weak supervision derived from the clinical annotation set $A_i$ and the class label $y_i$. Formally, the framework aims to learn a function $\mathcal{F}: \mathcal{P} \rightarrow \mathcal{A}$, where $\mathcal{P}$ is the space of image patches and $\mathcal{A}$ denotes the space of possible patch annotation labels.  
Each image $X_i \in \mathcal{D}$ is partitioned into non-overlapping $128 \times 128$ patches, reflecting the desired patch size, such that
\begin{equation}
    X_i = \{ r^i_j | j=1,\cdots,M_i; r_j^i \in \mathbb{R}^{128 \times 128 \times C}\},
\end{equation}
where $M_i$ denotes the number of patches extracted from the image $X_i$. For any two patches $r^i_{j_1}, r^i_{j_2} \in X_i$, $r^i_{j_1} \cap r^i_{j_2} = \emptyset$. 

Each extracted patch \( r_j^i \in X_i \) is assigned a preliminary label \( \ell(r_j^i) \) based on the supervision of the image-level class label $y_i$ and its overlap with the annotated regions in \( A_i \), represented as
\begin{equation}
\label{eq: preliminary label assignment}
    \ell(r_j^i) =
    \begin{cases}
        \texttt{Healthy}, & \text{if } y_i = \texttt{No-DR}, \\
        \texttt{Unhealthy}, & \text{if } y_i = \texttt{DR }\& \exists a^i_k \in A_i \text{ with } r_j^i \cap a^i_k \neq \emptyset,\\
        \texttt{Unlabeled}, & \text{if } y_i = \texttt{DR }\& \forall a^i_k \in A_i \text{ with } r_j^i \cap a^i_k = \emptyset.
    \end{cases}
\end{equation}
Here, unlabeled patches are defined as those extracted from images affected by DR without any overlap with any annotated lesion in $A_i$.
The complete set of unlabeled patches in all $X_i \in \mathcal{D}$ is indicated by $\mathcal{P}_\mathcal{U}$, whereas the complete set of labeled patches is given by $\mathcal{P}_\mathcal{L}$.
The total space of all image patches is defined as $\mathcal{P} = \mathcal{P}_\mathcal{L} \cup \mathcal{P}_\mathcal{U}$.

A key challenge comes from the coarse annotation of the lesion \( A_i \), as imprecise boundaries often include healthy neighboring pixels introducing label noise into the patch label set \( \mathcal{P}_\mathcal{L} \). For instance, if an annotation \( a^i_k \in A_i \) overlaps with an adjacent healthy patch \( r^i_{j} \), then \( r^i_{j} \cap a^i_k \ne \emptyset \) holds, causing \( r^i_{j} \) to be incorrectly labeled as \texttt{Unhealthy}.

\subsection{SAFE framework}

The Similarity-based Annotation via Feature-space Ensemble (SAFE) is a two-stage framework. First, it learns a structured representation space $(\mathcal{E})$ to embed patches based on their semantic similarity.  In the second stage, it learns to annotate patches in \( \mathcal{P}_\mathcal{U} \) using weak supervision from labeled patches in \( \mathcal{P}_\mathcal{L} \) and the structured patch representation space $(\mathcal{E})$ learned in the previous stage. Fig.~\ref{fig:SAFESchematic} illustrates the overall workflow of the framework.

\subsubsection{Stage 1: Patch embedding network (PEN)}
\label{sec: pen}

The PEN, illustrated in Fig. \ref{fig:stage1Schematic}, learns the structured latent representations $(\mathcal{E})$ of the labeled patches in $\mathcal{P}_\mathcal{L}$ through a dual-arm architecture. The design employs a shared encoder $f_\theta$ to simultaneously optimize class-discriminative features and semantically structured embeddings, ensuring both label fidelity and representation generalization. The integration of label supervision aligns embeddings with class-specific decision boundaries, introducing explicit semantic structure in the learned embedding space.

The first arm of the model performs binary classification using a classification head $c_\omega(\cdot)$, optimized with a binary cross-entropy loss $(\mathcal{L}_{\text{BCE}})$. It guides the model in learning discriminative characteristics, using supervision based on the preliminary label \( \ell(r^i_j) \) of the patches. Each input patch $\{(r^i_j,\ell(r^i_j)\} \in \mathcal{P}_\mathcal{L}$ is first encoded as \( z^i_j = f_\theta(r^i_j) \in \mathbb{R}^d \)  in the $d$-dimensional space. The embedding $z^i_j$ is subsequently fed to $c_{\omega}(\cdot)$.

The second arm maps the features generated from $f_{\theta}(\cdot)$ into a contrastive space through a projection head $g_\phi$, where a supervised contrastive loss \citep{khosla2020supervised} is optimized. The \( \mathcal{L}_{\text{SCL}} \) maximizes agreement between semantically similar patches, and separates dissimilar ones in the normalized contrastive space. This enforces semantic alignment while capturing the structure at the lesion level. Here, $z^i_j$ is projected as \( v^i_j = g_\phi(z^i_j) \in \mathbb{R}^{d'} \) in the contrastive space $d' \le d$. The projection head isolates the contrastive loss optimization from $\mathcal{E}$. It enables better alignment of similar embedding without degrading representational quality. An augmented version of \(r^i_j\), denoted as \(r^{i+}_j\), is also passed through the same encoder \( f_\theta \) and projection head \( g_\phi \). The agreement between the resulting embeddings in the contrastive space is maximized using $\mathcal{L}_{\text{SCL}}$. Positive and negative pairs are formed within each training batch based on their patch labels. The cosine similarity is then used to compute their alignment in the contrastive space. Given two embeddings \( v^i_j \) and \( v^k_l \) in the contrastive space, the cosine similarity is defined as $\text{sim}(v^i_j, v^k_l) = \frac{v^i_j \cdot v^k_l}{\|v^i_j\| \, \|v^k_l\|}$.
Consequently,
\begin{align}
\label{eq:SCLobjective}
\mathcal{L}_{\text{SCL}} &= \sum_{(i,j) \in \mathcal{B}} \frac{-1}{|P(i,j)|} \sum_{(k,l) \in P(i,j)} \nonumber \\
&\quad \cdot \log \frac{ \exp \left( \frac{\text{sim}(v^i_j, v^k_l)}{\kappa} \right) }{ \sum\limits_{(m,n) \in \mathcal{S}(i,j)} \mathbf{1}_{[(m,n) \ne (i,j)]} \exp \left( \frac{\text{sim}(v^i_j, v^m_n)}{\kappa} \right) }
\end{align}

where \( v^i_j \) is the anchor embedding of patch \( r^i_j \in \mathcal{P}_\mathcal{L} \)  and \( \mathcal{B} \) denotes the set of patch indices in the training batch. Here, \(P(i,j)\) and \( \mathcal{S}(i,j) \) represent the sets of positive pairs and all batch indices for the anchor \( v^i_j \), respectively. Moreover, \( \mathbf{1}_{[(m,n) \ne (i,j)]} \) is an indicator function that excludes the anchor, and \( \kappa \in \mathbb{R}^+ \) is the temperature scaling parameter. The $\mathcal{L}_{\text{BCE}}$ is defined as
\begin{equation}
\label{eq: bce_loss}
\small
\mathcal{L}_{\text{BCE}} = - \sum_{(i,j) \in \mathcal{B}} \left[ \ell(r^i_j) \log \hat{\ell(r^i_j)} + (1 - \ell(r^i_j)) \log (1 - \hat{\ell(r^i_j)}) \right],
\end{equation}
where $\hat{\ell(r^i_j)}$ is the predicted label.
The combined loss is defined as
\begin{equation}
\label{eq: combine_loss}
    \mathcal{L} = \mathcal{L}_{\text{SCL}} + \lambda \mathcal{L}_{\text{BCE}}, \quad \lambda \in [0,1].
\end{equation}
Here, the two loss functions complement each other to promote semantic clustering, with compactness within and separation between each pair of partitions. The $\lambda$ is used to control the relative importance of $\mathcal{L}_{\text{BCE}}$ with respect to $\mathcal{L}_{\text{SCL}}$. An empirical value of $\lambda = 0.3$ was chosen to allow the dominance of $\mathcal{L}_{\text{SCL}}$, while discouraging strict adherence to the noisy preliminary label in low-veracity training data.
Although the encoder backbone selected was ResNet18 \citep{resnet}, the proposed framework can be adapted to any standard backbone architecture. 
The output of \(f_{\theta}\), the feature embedding space $(\mathcal{E})$, is passed to Stage 2 after training, for similarity-based annotation. 

In order to improve robustness and stability, while reducing model-specific biases, a set of \( M_T \) independent patch embedding models are trained with different training folds. The encoder of the \( m \)th model instance, \( f^{(m)}_\theta \), produces a structured embedding space \( \mathcal{E}^{(m)} \). Given any patch \( r^i_j \in \mathcal{P}_\mathcal{L} \cup \mathcal{P}_\mathcal{U} \), the model generates its feature embedding \( \mathcal{E}^{(m)}(r^i_j) \in \mathbb{R}^d \). Sets of embeddings for labeled and unlabeled patches, for each model, are denoted as \( \mathcal{E}_L^{(m)} \) and \( \mathcal{E}_U^{(m)} \), respectively. Embeddings of all \( M_T \) models collectively form the basis for similarity-based annotation inference in Stage 2 of SAFE.

\subsubsection{Stage 2: Annotation via feature space ensemble}
\label{stage2}

This stage focuses on annotating the unlabeled patches using the structured embedding space $(\mathcal{E})$ constructed in Stage 1.
Since both $\mathcal{P_L}$ and $\mathcal{P_U}$ are drawn from the same dataset $\mathcal{D}$, they follow a common data distribution. Consequently, their embeddings remain aligned within each $\mathcal{E}^{(m)}$, facilitating reliable neighborhood-based inference across patch categories.
For each unlabeled patch \( r^p_q \in \mathcal{P_U} \), the framework computes a $d$-dimensional embedding \( \mathcal{E}^{(m)}(r^p_q) \in \mathcal{E}_U^{(m)} \), and compares it with all labeled embeddings \( \mathcal{E}_L^{(m)} \) using cosine similarity. The set of pairwise distances $\Delta_{pq}$ is the collection of distances $\delta_{(p,q)(i,j)}$ between $r^p_q$ and all $r^i_j \in \mathcal{E}_L^{(m)}$, where
\begin{equation}
\label{eq:cosine_distance}
    \delta_{(p,q)(i,j)} = 1 - \frac{\mathcal{E}^{(m)}(r^p_q) \cdot \mathcal{E}^{(m)}(r^i_j)}{\| \mathcal{E}^{(m)}(r^p_q) \| \cdot \| \mathcal{E}^{(m)}(r^i_j) \|}, \quad \forall r^i_j \in \mathcal{P_L}.
\end{equation}
All labeled patches are ranked, based on the ascending order $(\Delta_{pq}')$ of $\Delta_{pq}$. The top-\( K \) nearest neighbors \( (\mathcal{N}_K(r^p_q)) \) of $r^p_q$ are selected, followed by counting the number of \texttt{Healthy} and \texttt{Unhealthy} labels among these neighbors, \( n_H \) and \( n_U \), respectively. A tunable confidence threshold \( \tau \in [0.5, 1] \) defines the critical value 
\begin{equation}
\label{eq:criticalval}
    C_{\text{val}} = K \times \tau.
\end{equation}
Based on $C_{\text{val}}$, the model $m$ assigns a provisional label to the unlabeled patch as
\begin{equation}
    \ell^{(m)}(r^p_q) =
    \begin{cases}
        \texttt{Healthy}, & \text{if } n_H > C_{\text{val}}, \\
        \texttt{Unhealthy}, & \text{if } n_U > C_{\text{val}}, \\
        \texttt{Undecided}, & \text{otherwise}.
    \end{cases}
\end{equation}
The process is independently repeated, across all \( M_T \) trained patch embedding models, to generate a stable set of provisional annotations for each patch defined as 
\begin{equation}
    \mathcal{S}_{\text{prov}}(r^p_q) = \left\{ \ell^{(1)}(r^p_q), \ell^{(2)}(r^p_q), \cdots, \ell^{(M_T)}(r^p_q) \right\}.
\end{equation}
The majority vote strategy is then applied on \( \mathcal{S}_{\text{prov}}(r^p_q) \) to determine the final annotation \( \ell_{\text{pred}}(r^p_q) \). If no label receives a majority, the patch is assigned to the \texttt{Undecided} category -- enabling abstention in ambiguous cases. The complete procedure, which details the second stage of the SAFE framework, is described in Algorithm~\ref{alg:label_inference}. \hspace{2cm} $\square$

\begin{algorithm}[!th]
\footnotesize
\caption{Generating Annotations using Feature-space Ensemble}
\label{alg:label_inference}
\KwIn{
    \begin{itemize}
    \item Trained encoder–embedding space pairs \( \{(f_\theta^{(m)}, \mathcal{E}^{(m)})\}_{m=1}^{M_T} \)
    \item Unlabeled patch and embeddings \( (r^p_q, \mathcal{E}^{(m)}(r^p_q)) \)
    \item Labeled patch set and embeddings \( \{(\mathcal{P_L}, \mathcal{E}_L^{(m)})\}_{m=1}^{M_T} \),
    where \( \mathcal{E}_L^{(m)} = \left\{ \mathcal{E}^{(m)}(r^i_j) \mid r^i_j \in \mathcal{P_L} \right\} \)
    \item Number of nearest neighbors \( K \)
    \item Confidence threshold \( \tau \)
    \end{itemize}
}
\KwOut{
     Inferred annotation $\ell_{\text{pred}}(r^p_q)$
}

\textbf{Procedure:} \\
Initialize provisional annotation set $\mathcal{S}_{\text{prov}}(r^p_q) \gets \emptyset$.
    
    \ForEach{$m = 1$ to $M_T$}{
        Fetch feature embedding $\mathcal{E}^{(m)}(r^p_q)$ from \( \mathcal{E}^{(m)}_U\).\\
        Compute \( \delta_{(p,q)(i,j)} \) by Eqn. \eqref{eq:cosine_distance} for all \( \mathcal{E}^{(m)}(r^i_j) \in \mathcal{E}_L^{(m)} \), \allowbreak and collect the results in \( \Delta_{pq} \). \\
        Compute sorted distance set \( \Delta_{pq}' \gets \text{Sort}(\Delta_{pq}) \) in ascending order.\\
        Let \( \mathcal{N}_K(r^p_q) \) denote the top-\( K \) nearest labeled neighbors of \( r^p_q \), based on the sorted distance set \( \Delta_{pq}' \).\\
        Compute the number of \texttt{Healthy} and \texttt{Unhealthy} labels among the neighbors in \( \mathcal{N}_K(r^p_q) \) as
        \[
        n_H = \sum_{r \in \mathcal{N}_K(r^p_q)} \mathbf{1}_{[\ell(r) = \texttt{Healthy}]},
        \]\\
        \[
        n_U = \sum_{r \in \mathcal{N}_K(r^p_q)} \mathbf{1}_{[\ell(r) = \texttt{Unhealthy}]}.
        \]\\
        Compute critical value $C_{\text{val}} = K \times \tau$ by Eqn. (\ref{eq:criticalval}).
        
        \eIf{$n_H > C_{\text{val}}$}{
            Assign provisional annotation $\ell^{(m)}(r_p) \gets \texttt{Healthy}$ \\
        }{
            \eIf{$n_U > C_{\text{val}}$}{
                Assign provisional annotation $\ell^{(m)}(r_p) \gets \texttt{Unhealthy}$ \\
            }{
                Assign provisional annotation $\ell^{(m)}(r_p) \gets \texttt{Undecided}$. 
            }
        }
        
        Append provisional annotation  $\mathcal{S}_{\text{prov}}(r^p_q) \gets \mathcal{S}_{\text{prov}}(r^p_q) \cup \{\ell^{(m)}(r^p_q)\}$. 
    }
    
    Apply majority voting on $\mathcal{S}_{\text{prov}}(r^p_q)$ to infer final annotation $\ell_{\text{pred}}(r^p_q)$. \\
\Return $\ell_{\text{pred}}(r^p_q)$
\end{algorithm}

Thereby, the annotations generated by SAFE for previously unlabeled patches can be used to reconstruct fully annotated images as illustrated in Fig. \ref{fig:SAFESchematic}.

\subsubsection{Time complexity}
\label{subsec:timecomplexity}

Let $|\mathcal{P_U}|$ and $|\mathcal{P_L}|$ denote the number of unlabeled and labeled patches, respectively. For each unlabeled patch, computing the cosine similarity with all $|\mathcal{P_L}|$ labeled embeddings requires $O(d|\mathcal{P_L}|)$ operations. Identifying the top-$K$ neighbors, with sorting, adds an additional cost of \allowbreak
$O(|\mathcal{P_L}|  \log |\mathcal{P_L}|)$. Thus, the total complexity to generate annotation per patch, using Algorithm \ref{alg:label_inference}, becomes $ O(d  |\mathcal{P_L}| + |\mathcal{P_L}|  \log |\mathcal{P_L}|)$. Assuming $d$ to be a constant, this simplifies to $O(|\mathcal{P_L}| \log |\mathcal{P_L}|)$. 

\subsection{Datasets}
\label{sec:dataset}

The experiments were conducted on four datasets, {\it viz.}  Messidor \citep{messidor}, IDRiD \citep{idrid}, e-ophtha \citep{decenciere2013teleophta}, and DDR \citep{li2019diagnostic}. The latter three provided publicly available image-level labels as either \texttt{No-DR} or \texttt{DR}, along with the corresponding lesion annotations. For the Messidor dataset, with publicly available image labels, the lesion annotations were privately generated by the clinical expert in our team.

The framework used curated subsets of each dataset to ensure reliable lesion-level supervision and prevention of data leakage. The Messidor dataset ($N = 1200$), with annotations of the lesions, is hereby referred to as the Messidor$^*$ dataset. The IDRiD$^{(-)}$ dataset ($N = 249$) combined all samples affected by DR, with lesion masks from the segmentation partition, along with all No-DR images from the classification partition of the IDRiD dataset ($N = 516$). The e-ophtha$^{(-)}$ dataset ($N = 157$) retained one image per patient from the original e-ophtha dataset ($N = 463$) to avoid data leakage. This included all No-DR images, along with those having either exudate or microaneurysm annotations (marked DR-affected). The DDR$^{(-)}$ dataset ($N = 3263$) included DR-affected samples with lesion annotations, along with a randomly sampled balanced subset of No-DR images from the DDR dataset ($N = 13,673$).

The SAFE framework padded each image with black pixels so that the dimensions were divisible by the patch size. This enabled uniform patch extraction from images, with varying dimensions, in the different datasets. A patch size of $128 \times 128 \times 3$ was chosen to balance between contextual richness and computational efficiency. Patches containing more than 90\% black pixels were discarded. Table~\ref{tab:dataset_statistics} summarizes the details of the image samples and the number of patches extracted in all datasets. 

\begin{table}[!th]
\centering
\caption{Distribution of \texttt{No-DR} and \texttt{DR}-affected images in each dataset, along with the corresponding labeled and unlabeled patches.}
\label{tab:dataset_statistics}

\begingroup

\setlength{\tabcolsep}{6pt} 
\renewcommand{\arraystretch}{1.25} 
\scalebox{0.8}{
\begin{tabular}{|l|r|r|r|r|r|}
\hline
\multirow{3}{*}{\textbf{Dataset}} & \multicolumn{2}{c|}{\multirow{2}{*}{\textbf{Images}}} & \multicolumn{3}{c|}{\textbf{Patches}}                                                               \\ \cline{4-6} 
                         & \multicolumn{2}{c|}{}                        & \multicolumn{2}{c|}{\textbf{Labeled}}                                  & \multirow{2}{*}{\textbf{Unlabeled}} \\ \cline{2-5}
                         & \textbf{No-DR}       & \textbf{DR}        & \textbf{Healthy} & \textbf{Unhealthy} &                            \\ \hline
\textbf{Messidor$^*$}             & 546         & 654       & 41386   & 9387      & 39598                      \\ \hline
\textbf{IDRiD$^{(-)}$}                & 168         & 81        & 87987   & 9292      & 33490                      \\ \hline
\textbf{e-ophtha$^{(-)}$}             & 99          & 58        & 10522   & 511       & 8933                       \\ \hline
\textbf{DDR$^{(-)}$}                  & 2506        & 757       & 223542  & 18468     & 117448                     \\ \hline
\end{tabular}}
\endgroup%
\footnotesize
\\$*$ symbolizes original data with private annotation.
\\$^{(-)}$ symbolizes subset of the original dataset based on annotation availability.
\end{table}

\subsection{Experimental setup}
\label{sec:exp_setup}

\begin{figure}
    \centering
    \includegraphics[width=0.35\columnwidth]{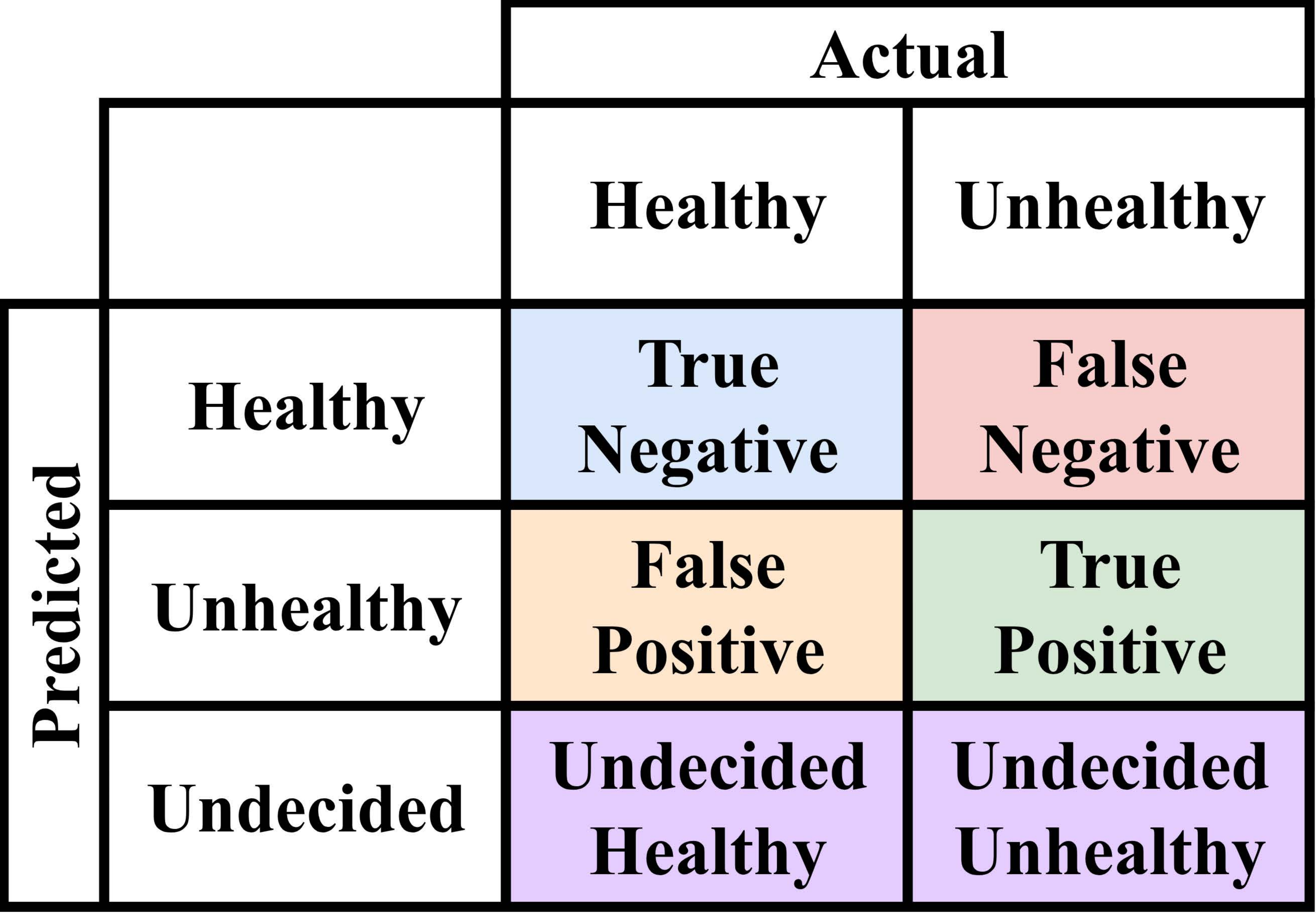}
    \caption{The extended confusion matrix used for computing metrics, such as Acc, BAcc, PR, RE, F1-score, AUPRC, D$_{\text{rate}}$ and MR}.
    \label{fig:cm}
\end{figure}

All experiments were conducted on a 12 GB NVIDIA Titan XP GPU. An 80-20 stratified split was performed on each dataset to create training and holdout test sets for the evaluation of the SAFE framework. A 4-fold stratified cross-validation was performed on the training dataset to generate $M_T = 3$ trained patch embedding networks (PEN) for ensuring ensemble diversity while preserving representative feature density across sparse folds. Each patch embedding model was trained for 150 epochs, using the Adam optimizer with a learning rate of $10^{-4}$ and a batch size of 128. Each model was trained with L1 and L2 regularization, both applied with a coefficient of $10^{-4}$. The values of both $d$ and $d^{\prime}$ were set to 512 in our experiments. Each patch embedding network comprised of 11.95M trainable parameters and executed approximately 0.60~GFLOPs for an $128 \times 128 \times 3$ input image.

Each mini-batch was stratified to support class diversity in each training batch while optimizing $\mathcal{L}_{\text{SCL}}$. The temperature $\kappa$ was set to 0.035. Data augmentation included horizontal and vertical flips, random rotation, vignetting, grayscale conversion, shearing, Gaussian blur, and color distortion.
Exhaustive search over all $K$–$\tau$ combinations was computationally infeasible. Therefore, guided empirical tuning (of Section \ref{subsec:senanal}) determined the number of nearest neighbors $K$ and the confidence threshold $\tau$. The reported values represent configurations that produced stable and consistent performance across the datasets.

The evaluation of the SAFE framework used a held-out test set, which was segregated into labeled and unlabeled patches following the strategy of Section~\ref{sec: problemformulation}. The labeled subset served as ground truth for comparing the annotations predicted by SAFE. Note that the inclusion of the \texttt{Undecided} category in the SAFE framework redirected uncertain predictions to a separate class, resulting in the formation of a nonstandard confusion matrix (as shown in Fig. \ref{fig:cm}). The effectiveness of SAFE was demonstrated through annotation of unlabeled patches, which were then evaluated in a downstream classification task.

\subsection{Evaluation metrics}
\label{subsec:evalMethod}

The Davies-Bouldin (DB) index \citep{davies2009cluster} quantifies intra-class compactness and inter-class separation in the embedding space. The Wasserstein Distance (WD) \citep{shen2018wasserstein} measures the distributional alignment between the seen and unseen data. Both metrics range from 0 to $\infty$, with lower values indicating better performance. They evaluated the impact of the combined loss formulation and the effect of varying $\lambda$ in $\mathcal{L}$ in the SAFE framework.
Accuracy (Acc), Balanced Accuracy (BAcc), Precision (PR), Recall (RE), F1-score, and AUPRC \citep{pedregosa2011scikit} assessed the annotation quality of SAFE against expert annotations serving as ground truth. The PR, RE and F1-score captured the class-wise performance. 

We introduced the Decided Rate (D$_{\text{rate}}$) and extended Misclassification Rate ($MR$) to evaluate the proportions of the total annotated and correctly annotated samples generated by SAFE.
The Decided Rate (D$_{\text{rate}}$) is defined as the ratio of total inferred \texttt{Healthy} and \texttt{Unhealthy} samples to the total number of samples to be annotated. The metric, which varies between 0 and 1, provides insight into the annotation coverage of the model; with a high D$_{\text{rate}}$ being desirable.
The extended Misclassification Rate (MR) for a given class is defined as the ratio of samples from that class that are incorrectly classified to the total number of actual samples in that class. Unlike other metrics, $MR$ takes the samples in \texttt{Undecided} category in account. Eqns. (1)-(2) in the Supplementary Material provide their formulation.
For all evaluation metrics, with the exception of DB, WD, and MR, a higher value was indicative of better performance.

\section{Experimental Results and Discussion}
\label{sec:experimentaleval}

A detailed evaluation of the SAFE framework is presented, using quantitative metrics and qualitative analysis, for ablation studies as well as comparison with related state-of-the-art methods. The effect of individual loss functions and the variation of $\lambda$ in $\mathcal{L}$ is analyzed along with qualitative insights through activation maps and Gradient-weighted Class Activation Map (Grad-CAM) \citep{cao2022collaborative} visualization. In all tables {\bf bold} indicates best values and {\it italics} represents second best. Note that all analyses are performed at the patch level, except for the qualitative validation of annotations generated by SAFE in Fig. \ref{fig:annotation_full} of Section \ref{subsec:annotation_validation}. 

\subsection{Comparative study}
\label{subsec:ComAna}

The SAFE framework was compared with five strong baselines, as summarized in Table~\ref{tab:baselineCom} on the four datasets used. All experiments strictly followed the implementation strategy provided in the corresponding publications. \\ (1) Vanilla ResNet18 classifier \citep{resnet} -- weakly supervised label inference strategy of SAFE vs fully supervised training, when patch labels were scarce and noisy;\\ (2) Lesion-based Contrastive Learning (LCL) \citep{huang2020automated} for the detection of DR; \\ (3) PEN with traditional K-NN classifier (PEN + KNN) -- assessing the use of a higher threshold $\tau$ in SAFE over the standard K-NN-based annotation; \\ (4) Prototype-based Label Transfer (PLT) \citep{snell2017prototypical} -- contrasting with single-prototype matching in representation space; and \\  (5) Deep Cluster \citep{caron2018deep} -- unsupervised iterative learning of features while clustering data without labels.

SAFE consistently achieved the highest overall Acc and F1-score, along with the lowest MR on all four datasets. It rarely misclassified \texttt{Healthy} patches as \texttt{Unhealthy}, with a high precision for {\it Unhealthy}. A high \texttt{Healthy} precision implied that it avoided labeling other classes as \texttt{Healthy}. At the same time, the lower recall of \texttt{Unhealthy} indicated failure to identify some of the true \texttt{Unhealthy} patches. This suggests that, instead of making incorrect predictions, SAFE abstained from labeling uncertain and \texttt{Unhealthy} patches but assigned them to the \texttt{Undecided} class. Besides, its high \texttt{Healthy} Recall confirms that it correctly identified most actual \texttt{Healthy} patches.
The data-specific analysis is as follows.

\noindent \textit{{\bf Messidor$^*$}}: SAFE ($K=25, \tau=0.75$) achieved the highest scores in almost all evaluation metrics. The highest precision and recall values in the \texttt{Healthy} category indicate that SAFE accurately identified and correctly annotated almost all \texttt{Healthy} patches. The highest precision in the \texttt{Unhealthy} category confirms minimal mislabeling of \texttt{Healthy} as \texttt{Unhealthy}. Although its recall in \texttt{Unhealthy} was second-best, the combination of high F1-score in \texttt{Healthy} and strong abstention behavior suggest that SAFE assigned uncertain \texttt{Unhealthy} cases to the \texttt{Undecided} class rather than making incorrect predictions. This corroborates the reliability of SAFE in annotating the rare class(es). Although LCL marginally improved recall for \texttt{Unhealthy}, only SAFE achieved the lowest MR in both classes; thereby, reinforcing its quality of annotation.

\begin{table*}[!t]
\centering
\caption{Comparative analysis of SAFE with baseline approaches.}
\label{tab:baselineCom}

\begingroup

\setlength{\tabcolsep}{6pt} 
\renewcommand{\arraystretch}{1.25} 
\scalebox{0.75}{
\begin{tabular}{|l|l|r|r|rrrr|rrrr|r|}
\hline
 \multirow{2}{*}{\textbf{Dataset}} & \multirow{2}{*}{\textbf{Method}} & \multicolumn{1}{l|}{\multirow{2}{*}{\textbf{Acc ($\uparrow$)}}} & \multicolumn{1}{l|}{\multirow{2}{*}{\textbf{BAcc ($\uparrow$)}}} & \multicolumn{4}{c|}{\textbf{Healthy}}                                                                                                                                                           & \multicolumn{4}{c|}{\textbf{Unhealthy}}                                                                                                                                                         & \multicolumn{1}{l|}{\multirow{2}{*}{\textbf{$\mathbf{D}_{\text{rate}}$~($\uparrow$)}}} \\ \cline{5-12}
                                  &                                  & \multicolumn{1}{l|}{}                                           & \multicolumn{1}{l|}{}                                            & \multicolumn{1}{c|}{\textbf{PR ($\uparrow$)}} & \multicolumn{1}{c|}{\textbf{RE ($\uparrow$)}} & \multicolumn{1}{c|}{\textbf{F1 ($\uparrow$)}} & \multicolumn{1}{c|}{\textbf{MR ($\downarrow$)}} & \multicolumn{1}{c|}{\textbf{PR ($\uparrow$)}} & \multicolumn{1}{c|}{\textbf{RE ($\uparrow$)}} & \multicolumn{1}{c|}{\textbf{F1 ($\uparrow$)}} & \multicolumn{1}{c|}{\textbf{MR ($\downarrow$)}} & \multicolumn{1}{l|}{}                                                                  \\ \hline
\multirow{6}{*}{\textbf{Messidor$^*$}} & \textbf{SAFE} & \textbf{0.9599} & \textbf{0.9062} & \multicolumn{1}{r|}{\textbf{0.9601}} & \multicolumn{1}{r|}{\textbf{0.9919}} & \multicolumn{1}{r|}{\textbf{0.9758}} & \textbf{0.0077} & \multicolumn{1}{r|}{\textbf{0.9589}} & \multicolumn{1}{r|}{{\textit{ 0.8205}}} & \multicolumn{1}{r|}{\textbf{0.8844}} & \textbf{0.1529} & {\textit {0.9375}} \\
 & LCL & 0.9347 & {\textit {0.8932}} & \multicolumn{1}{r|}{{\textit {0.9546}}} & \multicolumn{1}{r|}{0.9638} & \multicolumn{1}{r|}{0.9592} & 0.0362 & \multicolumn{1}{r|}{0.8546} & \multicolumn{1}{r|}{\textbf{0.8227}} & \multicolumn{1}{r|}{0.8384} & {\textit {0.1773}} & \textbf{1.0000} \\
 & ResNet18 & 0.9238 & 0.8853 & \multicolumn{1}{r|}{0.9533} & \multicolumn{1}{r|}{0.9507} & \multicolumn{1}{r|}{0.9520} & 0.0493 & \multicolumn{1}{r|}{0.8114} & \multicolumn{1}{r|}{0.8200} & \multicolumn{1}{r|}{0.8157} & 0.1800 & \textbf{1.0000} \\
 & PEN + KNN & {\textit {0.9460}} & 0.8881 & \multicolumn{1}{r|}{0.9495} & \multicolumn{1}{r|}{{\textit {0.9849}}} & \multicolumn{1}{r|}{{\textit {0.9668}}} & {\textit {0.0150}} & \multicolumn{1}{r|}{{\textit {0.9292}}} & \multicolumn{1}{r|}{0.7914} & \multicolumn{1}{r|}{{\textit {0.8548}}} & 0.2007 & \textbf{1.0000} \\
  & PLT & 0.7765 & 0.7037 & \multicolumn{1}{r|}{0.8800} & \multicolumn{1}{r|}{0.8300} & \multicolumn{1}{r|}{0.8500} & 0.1700 & \multicolumn{1}{r|}{0.4700} & \multicolumn{1}{r|}{0.5800} & \multicolumn{1}{r|}{0.5200} & 0.4200 & \textbf{1.0000} \\
 & DeepCluster & 0.4028 & 0.5540 & \multicolumn{1}{r|}{0.8586} & \multicolumn{1}{r|}{0.2972} & \multicolumn{1}{r|}{0.4416} & 0.7028 & \multicolumn{1}{r|}{0.2299} & \multicolumn{1}{r|}{0.8108} & \multicolumn{1}{r|}{0.3582} & 0.1892 & \textbf{1.0000} \\ \hline
\multirow{6}{*}{\textbf{IDRiD$^{(-)}$}} & \textbf{SAFE} & \textbf{0.9739} & \textbf{0.8637} & \multicolumn{1}{r|}{\textbf{0.9767}} & \multicolumn{1}{r|}{\textbf{0.9954}} & \multicolumn{1}{r|}{\textbf{0.9859}} & \textbf{0.0045} & \multicolumn{1}{r|}{\textbf{0.9335}} & \multicolumn{1}{r|}{{\textit {0.7320}}} & \multicolumn{1}{r|}{\textbf{0.8205}} & \textbf{0.2158} & {\textit {0.9568}} \\
 & LCL & 0.9577 & {\textit {0.8632}} & \multicolumn{1}{r|}{{\textit {0.9730}}} & \multicolumn{1}{r|}{0.9805} & \multicolumn{1}{r|}{0.9767} & 0.0195 & \multicolumn{1}{r|}{0.8039} & \multicolumn{1}{r|}{\textbf{0.7460}} & \multicolumn{1}{r|}{{\textit {0.7739}}} & {\textit {0.2540}} & \textbf{1.0000} \\
 & ResNet18 & 0.9554 & 0.8547 & \multicolumn{1}{r|}{0.9713} & \multicolumn{1}{r|}{0.9796} & \multicolumn{1}{r|}{0.9754} & 0.0204 & \multicolumn{1}{r|}{0.7934} & \multicolumn{1}{r|}{0.7298} & \multicolumn{1}{r|}{0.7603} & 0.2702 & \textbf{1.0000} \\
 & PEN + KNN & {\textit {0.9593}} & 0.8266 & \multicolumn{1}{r|}{0.9647} & \multicolumn{1}{r|}{{\textit {0.9912}}} & \multicolumn{1}{r|}{{\textit {0.9778}}} & {\textit {0.0088}} & \multicolumn{1}{r|}{{\textit {0.8894}}} & \multicolumn{1}{r|}{0.6620} & \multicolumn{1}{r|}{0.7590} & 0.3380 & \textbf{1.0000} \\
  & PLT & 0.7141 & 0.6320 & \multicolumn{1}{r|}{0.9357} & \multicolumn{1}{r|}{0.7338} & \multicolumn{1}{r|}{0.8226} & 0.2662 & \multicolumn{1}{r|}{0.1761} & \multicolumn{1}{r|}{0.5301} & \multicolumn{1}{r|}{0.2643} & 0.4699 & \textbf{1.0000} \\
 & DeepCluster & 0.3600 & 0.4814 & \multicolumn{1}{r|}{0.8934} & \multicolumn{1}{r|}{0.3309} & \multicolumn{1}{r|}{0.4829} & 0.6691 & \multicolumn{1}{r|}{0.0920} & \multicolumn{1}{r|}{0.6319} & \multicolumn{1}{r|}{0.1606} & 0.3681 & \textbf{1.0000} \\ \hline
\multirow{6}{*}{\textbf{e-ophtha$^{(-)}$}} & \textbf{SAFE} & \textbf{0.9829} & 0.7729 & \multicolumn{1}{r|}{\textbf{0.9837}} & \multicolumn{1}{r|}{\textbf{0.9989}} & \multicolumn{1}{r|}{\textbf{0.9912}} & \textbf{0.0011} & \multicolumn{1}{r|}{\textbf{0.9459}} & \multicolumn{1}{r|}{0.5469} & \multicolumn{1}{r|}{\textbf{0.6931}} & \textbf{0.2685} & {\textit {0.9557}} \\
 & LCL & 0.9620 & {\textit {0.7884}} & \multicolumn{1}{r|}{0.9756} & \multicolumn{1}{r|}{0.9843} & \multicolumn{1}{r|}{0.9800} & 0.0157 & \multicolumn{1}{r|}{0.6957} & \multicolumn{1}{r|}{{\textit {0.5926}}} & \multicolumn{1}{r|}{{\textit {0.6400}}} & 0.4074 & \textbf{1.0000} \\
 & ResNet18 & 0.9473 & \textbf{0.8198} & \multicolumn{1}{r|}{{\textit {0.9801}}} & \multicolumn{1}{r|}{0.9637} & \multicolumn{1}{r|}{0.9718} & 0.0363 & \multicolumn{1}{r|}{0.5290} & \multicolumn{1}{r|}{\textbf{0.6759}} & \multicolumn{1}{r|}{0.5935} & {\textit {0.3241}} & \textbf{1.0000} \\
 & PEN + KNN & {\textit {0.9631}} & 0.7151 & \multicolumn{1}{r|}{0.9669} & \multicolumn{1}{r|}{{\textit {0.9950}}} & \multicolumn{1}{r|}{{\textit {0.9807}}} & {\textit {0.0050}} & \multicolumn{1}{r|}{{\textit {0.8393}}} & \multicolumn{1}{r|}{0.4352} & \multicolumn{1}{r|}{0.5732} & 0.5648 & \textbf{1.0000} \\
  & PLT & 0.7026 & 0.6379 & \multicolumn{1}{r|}{0.9600} & \multicolumn{1}{r|}{0.7100} & \multicolumn{1}{r|}{0.8200} & 0.2662 & \multicolumn{1}{r|}{0.1100} & \multicolumn{1}{r|}{0.5600} & \multicolumn{1}{r|}{0.1800} & 0.4699 & \textbf{1.0000} \\
 & DeepCluster & 0.6964 & 0.4214 & \multicolumn{1}{r|}{0.9317} & \multicolumn{1}{r|}{0.7317} & \multicolumn{1}{r|}{0.8197} & 0.2683 & \multicolumn{1}{r|}{0.2440} & \multicolumn{1}{r|}{0.1111} & \multicolumn{1}{r|}{0.0400} & 0.8889 & \textbf{1.0000} \\ \hline
\multirow{6}{*}{\textbf{DDR$^{(-)}$}} & \textbf{SAFE} & \textbf{0.9886} & {\textit {0.9209}} & \multicolumn{1}{r|}{{\textit {0.9893}}} & \multicolumn{1}{r|}{\textbf{0.9986}} & \multicolumn{1}{r|}{\textbf{0.9939}} & \textbf{0.0014} & \multicolumn{1}{r|}{\textbf{0.9768}} & \multicolumn{1}{r|}{{\textit {0.8431}}} & \multicolumn{1}{r|}{\textbf{0.9051}} & {\textit {0.1310}} & {\textit {0.9779}} \\
 & LCL & 0.9779 & 0.8875 & \multicolumn{1}{r|}{0.9823} & \multicolumn{1}{r|}{0.9941} & \multicolumn{1}{r|}{0.9881} & 0.0059 & \multicolumn{1}{r|}{0.9152} & \multicolumn{1}{r|}{0.7809} & \multicolumn{1}{r|}{0.8427} & 0.2191 & \textbf{1.0000} \\
 & ResNet18 & 0.9564 & \textbf{0.9215} & \multicolumn{1}{r|}{\textbf{0.9899}} & \multicolumn{1}{r|}{0.9627} & \multicolumn{1}{r|}{0.9761} & 0.0373 & \multicolumn{1}{r|}{0.6588} & \multicolumn{1}{r|}{\textbf{0.8803}} & \multicolumn{1}{r|}{0.7536} & \textbf{0.1197} & \textbf{1.0000} \\
 & PEN + KNN & {\textit {0.9794}} & 0.8795 & \multicolumn{1}{r|}{0.9808} & \multicolumn{1}{r|}{{\textit {0.9973}}} & \multicolumn{1}{r|}{{\textit {0.9890}}} & {\textit {0.0027}} & \multicolumn{1}{r|}{{\textit {0.9579}}} & \multicolumn{1}{r|}{0.7618} & \multicolumn{1}{r|}{{\textit {0.8486}}} & 0.2382 & \textbf{1.0000} \\
  & PLT & 0.7357 & 0.6502 & \multicolumn{1}{r|}{0.9393} & \multicolumn{1}{r|}{0.7562} & \multicolumn{1}{r|}{0.8379} & 0.2438 & \multicolumn{1}{r|}{0.1932} & \multicolumn{1}{r|}{0.5441} & \multicolumn{1}{r|}{0.2852} & 0.4559 & \textbf{1.0000} \\
 & DeepCluster & 0.6590 & 0.4456 & \multicolumn{1}{r|}{0.9135} & \multicolumn{1}{r|}{0.6971} & \multicolumn{1}{r|}{0.7908} & 0.3029 & \multicolumn{1}{r|}{0.0499} & \multicolumn{1}{r|}{0.1941} & \multicolumn{1}{r|}{0.0794} & 0.8059 & \textbf{1.0000}
 \\ \hline
\end{tabular}}
\endgroup%
\end{table*}

\noindent \textit{{\bf IDRiD$^{(-)}$}}: IDRiD shares common characteristics with Messidor, including sample size and class imbalance. SAFE ($K=25, \tau=0.70$) maintained similar performance trends here, achieving the highest Accuracy and F1-scores in both classes. The high annotation coverage, as indicated by $\text{D}_{\text{rate}}$, validates the consistency of its annotations in moderately sized datasets.

\noindent \textit{{\bf e-ophtha$^{(-)}$}}: e-ophtha$^{(-)}$ poses the greatest challenge, as it is the smallest and most imbalanced dataset. SAFE ($K=7, \tau=0.80$) achieved the best performance in all metrics for the \texttt{Healthy} class and attained the highest precision for \texttt{Unhealthy}, indicating its strong reliability. The moderate recall of \texttt{Unhealthy} reflects the cautious strategy of SAFE in low-data settings, where it abstained from uncertain predictions and assigned these to the \texttt{Undecided} class. Despite under-identifying some actual \texttt{Unhealthy} patches, SAFE avoided noisy or spurious labels; a limitation observed in others. This made SAFE more effective in data-scarce medical imaging scenario.

\noindent \textit{{\bf DDR$^{(-)}$}}: The large sample size and strong class imbalance in DDR offer a different perspective in evaluating SAFE. With $K=35$, $\tau=0.70$, it achieved the best or second-best performance on all metrics. Although a slightly lower \texttt{Unhealthy} recall marginally reduced BAcc, its high precision in both classes highlights the reliability of its annotation. This confirms that SAFE maintained its advantage even in large-scale settings; thereby, demonstrating its excellent scalability.

SAFE served as an effective weakly supervised annotation framework, particularly in class-imbalanced, noisy, or large-data scenarios. It consistently delivered good performance on all metrics, demonstrating strong scalability and good annotation quality with significantly high $\text{D}_{\text{rate}}$ $(>93\%)$. The threshold $\tau$ ensured high-precision annotation. Considering only the top $K$ neighbors, enabled context-sensitive and noise-resistant labeling. It is evident that SAFE incurred an additional computational cost in Stage 2 due to annotation propagation. However, the use of structured embeddings and a similarity-based strategy provided a clear advantage over supervised classifiers, K-NN-based methods, and unsupervised techniques. Compared to single prototype-based label transfer in PLT, our SAFE clearly outperformed. Its locality-based annotation, grounded in the neighborhood context, yielded reliable results. 

\subsection{Ablations}
\label{subsec:ablation}

A detailed analysis of the design choices under-pinning the SAFE framework is provided here. We investigate the impact of the ensemble strategy and loss function combination, on the reliability of annotations and embedding space characteristics.

\paragraph{Role of ensemble} Table \ref{tab:ablation_en} provides the performance analysis of SAFE without ensemble ({\it ie.}, one instance of PEN), and with ensemble ({\it ie.}, three instances of PEN). Without ensemble SAFE is found to achieve a marginally higher $D_{rate}$, resulting in consistently lower scores across all metrics (such as BAcc and F1-score) for all datasets. In contrast, the ensemble-based SAFE shows expected increase in both BAcc and F1-score for all datasets, with a marginal decrease in $D_{rate}$. This indicates the ensemble approach could successfully eliminate noisy label propagation, and make the model robust with reduced variability.

\begin{table}[!ht]
\centering
\caption{Comparative performance of SAFE variants, without and with ensemble.}
\label{tab:ablation_en}

\begingroup

\setlength{\tabcolsep}{6pt} 
\renewcommand{\arraystretch}{1.25} 
\scalebox{0.8}{
\begin{tabular}{|l|l|c|c|c|c|}
\hline
\textbf{Dataset}                   & \textbf{SAFE}    & \textbf{BAcc ($\uparrow$)} & \textbf{F1-Healthy ($\uparrow$)} & \textbf{F1-Unhealthy ($\uparrow$)} & \textbf{D$_{\text{rate}}$ ($\uparrow$)} \\ \hline
\multirow{2}{*}{\textbf{Messidor}} & without Ensemble & 0.8745                     & 0.9582                           & 0.8421                             & \textbf{0.9754}             \\ \cline{2-6} 
                                   & Ensemble         & \textbf{0.9062}            & \textbf{0.9758}                  & \textbf{0.8844}                    & 0.9375                      \\ \hline
\multirow{2}{*}{\textbf{IDRiD}}    & without Ensemble & 0.8211                     & 0.9674                           & 0.7812                             & \textbf{0.9882}             \\ \cline{2-6} 
                                   & Ensemble         & \textbf{0.8637}            & \textbf{0.9859}                  & \textbf{0.8205}                    & 0.9568                      \\ \hline
\multirow{2}{*}{\textbf{e-ophtha}} & without Ensemble & 0.7358                     & 0.9721                           & 0.6485                             & \textbf{0.9821}             \\ \cline{2-6} 
                                   & Ensemble         & \textbf{0.7729}            & \textbf{0.9912}                  & \textbf{0.6931}                    & 0.9557                      \\ \hline
\multirow{2}{*}{\textbf{DDR}}      & without Ensemble & 0.8894                     & 0.981                            & 0.8647                             & \textbf{0.9943}             \\ \cline{2-6} 
                                   & Ensemble         & \textbf{0.9209}            & \textbf{0.9939}                  & \textbf{0.9051}                    & 0.9779                      \\ \hline
\end{tabular}}
\endgroup%
\end{table}

\paragraph{Study on loss function} Table~\ref{tab:ablation} presents a comparison of the effects of optimizing PEN using loss functions $\mathcal{L}_{BCE}$, $\mathcal{L}_{SCL}$ and their combination $\mathcal{L}$ [of eqns. (\ref{eq:SCLobjective})-(\ref{eq: combine_loss})], on the quality of the learned embedding space in the SAFE framework. PEN aims to construct an embedding space with intra-cluster compactness, inter-cluster separation, and distributional alignment between the training and inference data. These constitute the three necessary conditions for reliable annotation transfer in SAFE.

\begin{table}[ht]
\centering
\caption{Comparative performance of PEN variants, optimized with loss functions \( \mathcal{L}_{\text{BCE}} \), \( \mathcal{L}_{\text{SCL}} \), and their combination \( \mathcal{L} \).}
\label{tab:ablation}

\begingroup

\setlength{\tabcolsep}{6pt} 
\renewcommand{\arraystretch}{1.25} 
\scalebox{0.8}{
\begin{tabular}{|l|l|c|cc|}
\hline
\multirow{2}{*}{\textbf{Dataset}} & \multirow{2}{*}{\textbf{\begin{tabular}[c]{@{}l@{}}Loss\\ Optimized\end{tabular}}} & \multirow{2}{*}{\textbf{DB}} & \multicolumn{2}{c|}{\textbf{WD}} \\ \cline{4-5} 
 &  &  & \multicolumn{1}{c|}{\textbf{Healthy}} & \textbf{Unhealthy} \\ \hline
\multirow{3}{*}{\textbf{Messidor$^{*}$}} & $\mathcal{L}_{BCE}$ & \textbf{0.8695 ± 0.1068} & \multicolumn{1}{c|}{0.0130 ± 0.0099} & 0.0563 ± 0.0384 \\
 & $\mathcal{L}_{SCL}$ & 0.9160 ± 0.0359 & \multicolumn{1}{c|}{\textit{0.0014 ± 0.0006}} & \textit{0.0075 ± 0.0045} \\
 & $\mathcal{L}$ & \textit{0.9096 ± 0.0130} & \multicolumn{1}{c|}{\textbf{0.0013 ± 0.0004}} & \textbf{0.0036 ± 0.0028} \\ \hline
\multirow{3}{*}{\textbf{IDRiD$^{(-)}$}} & $\mathcal{L}_{BCE}$ & \textbf{0.8485 ± 0.0087} & \multicolumn{1}{c|}{0.0042 ± 0.0009} & 0.1467 ± 0.0241 \\
 & $\mathcal{L}_{SCL}$ & 1.6056 ± 0.6535 & \multicolumn{1}{c|}{\textbf{0.0004 ± 0.0002}} & \textbf{0.0078 ± 0.0071} \\
 & $\mathcal{L}$ & \textit{0.8853 ± 0.0437} & \multicolumn{1}{c|}{\textit{0.0009 ± 0.0007}} & \textit{0.0166 ± 0.0061} \\ \hline
\multirow{3}{*}{\textbf{e-ophtha$^{(-)}$}} & $\mathcal{L}_{BCE}$ & 1.3959 ± 0.1010 & \multicolumn{1}{c|}{0.0216 ± 0.0133} & 0.2507 ± 0.1349 \\
 & $\mathcal{L}_{SCL}$ & \textit{1.2061 ± 0.0614} & \multicolumn{1}{c|}{\textit{0.0181 ± 0.0121}} & \textit{0.0413 ± 0.0058} \\
 & $\mathcal{L}$ & \textbf{1.1653 ± 0.1323} & \multicolumn{1}{c|}{\textbf{0.0031 ± 0.0015}} & \textbf{0.0263 ± 0.0121} \\ \hline
\multirow{3}{*}{\textbf{DDR$^{(-)}$}} & $\mathcal{L}_{BCE}$ & \textit{0.7294 ± 0.0212} & \multicolumn{1}{c|}{0.0023 ± 0.0021} & 0.1414 ± 0.1295 \\
 & $\mathcal{L}_{SCL}$ & 0.7849 ± 0.0351 & \multicolumn{1}{c|}{\textit{0.0003 ± 0.0007}} & \textit{0.0094 ± 0.0039} \\
 & $\mathcal{L}$ & \textbf{0.5834 ± 0.0168} & \multicolumn{1}{c|}{\textbf{0.0001 ± 0.0000}} & \textbf{0.0063 ± 0.0017} \\ \hline
 \multicolumn{5}{@{}l}{Values represent mean ± std in all cases.}
\end{tabular}}
\endgroup%
\end{table}

The study explored the contribution of each loss function to the structure and generalizability of the embedding space. 
Optimization with \( \mathcal{L}_{\text{BCE}} \) yielded a low DB index for larger datasets, indicating better clustering. However, the high WD score revealed poor generalization to unseen data, particularly for the \texttt{Unhealthy} class. This limitation became more pronounced in smaller datasets, such as e-ophtha, where it led to poor performance with both the DB index and the WD score.
In contrast, \( \mathcal{L}_{\text{SCL}} \) significantly improved WD, especially for the \texttt{Unhealthy} class. This indicated a stronger semantic understanding of abnormal regions. However, the relatively high DB-index score implied the inability of PEN to produce compact clusters with larger datasets. The low standard deviation, associated with \( \mathcal{L}_{\text{SCL}} \), pointed to a more consistent and stable embedding quality in all instances of PEN. 

The combined loss \( \mathcal{L} \) balanced both the necessary objectives. It reduced the DB-index more than \( \mathcal{L}_{\text{SCL}} \), and achieved significantly lower WD than \( \mathcal{L}_{\text{BCE}} \). This led to a desirable trade-off between cluster compactness and generalization. The combination achieved the best overall performance in the e-ophtha$^{(-)}$ and DDR$^{(-)}$ datasets, and was second in the remaining two (with only a marginal drop).

\begin{figure*}[!ht]
    \centering
    \includegraphics[width=0.85\columnwidth]{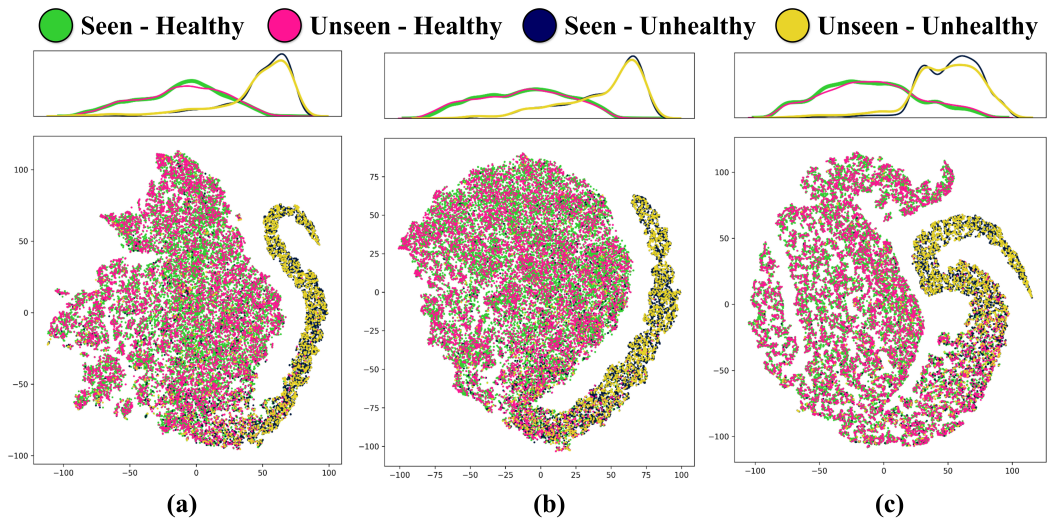}
    \caption{t-SNE embeddings of labeled training (seen) and test (unseen) patches from a Messidor split, obtained using PEN optimized with (a) \( \mathcal{L}_{\text{BCE}} \), (b) \( \mathcal{L}_{\text{SCL}} \), (c)  \( \mathcal{L} \).  }
    \label{fig:ablationvisual}
\end{figure*}

The t-SNE embeddings in Fig.~\ref{fig:ablationvisual} reinforce the trends observed in the quantitative results of Table~\ref{tab:ablation}. Optimization with \( \mathcal{L}_{\text{BCE}} \) produced compact clusters but did not maintain alignment between seen and unseen distributions. This caused the misplacement of a significant number of \texttt{Unhealthy} test cases (represented by ``\ybullet") within the \texttt{Healthy} cluster. In part (a), the distribution of the Unhealthy class in the seen data exhibits two distinct peaks; a pattern that is not observed in the unseen data. In contrast, \( \mathcal{L}_{\text{SCL}} \) improved alignment between ``seen" and ``unseen" data, with degraded cluster compactness. The combined loss $\mathcal{L}$ achieved a trade-off between structured clustering and distributional alignment, corroborating its balanced contribution to the SAFE framework.

\begin{figure*}[!ht]
    \centering
    \includegraphics[width=0.8\textwidth]{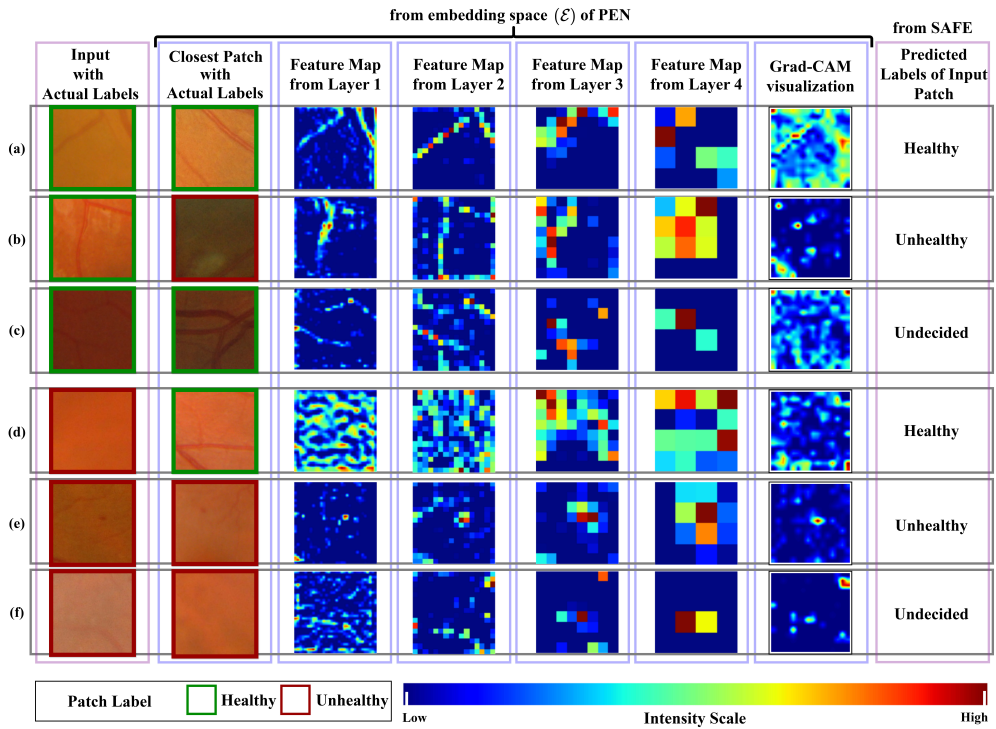}
    \caption{Visualization of patch-level explanation. Rows (a)–(c) are initially labeled patches \texttt{Healthy}, and (d)–(f) are initially labeled patches \texttt{Unhealthy}.}
    \label{fig:explaination}
\end{figure*}

\subsection{Visual interpretation}
\label{subsec:visualinter}

Fig.~\ref{fig:explaination} provides a representative qualitative explanation of the PEN model $f_{\theta}$, with embedding space $\mathcal{E}$. 
The examples, grouped by actual labels and predicted patch labels generated by SAFE, illustrate its semantic understanding of pathological features along with annotation behavior.

In Row (a) of the figure, the feature maps highlight vasculature in a true \texttt{Healthy} patch. This indicates that the model could associate the absence of lesions near vessels with confident \texttt{Healthy} predictions. Grad-CAM visualization shows diffuse attention throughout the patch, with a focus on vessels. This suggests that the model considered the entire patch, particularly vascular structures, before assigning a \texttt{Healthy} label. Row (b) presents an initially \texttt{Healthy}-labeled patch that the model subsequently inferred as \texttt{Unhealthy} by correctly identifying the subtle marks of lesion. The model embedded this patch near a true \texttt{Unhealthy} sample in the $\mathcal{E}$ obtained by PEN. Here the Grad-CAM view demonstrated that stage 1 of SAFE could accurately localize the abnormalities. This shows that the model could accurately detect the lesions, irrespective of the preliminary patch label marked Healthy. Row (c) shows a \texttt{Healthy}-labeled patch with poor contrast and inadequate visual features. Although it was embedded close to other \texttt{Healthy} samples in $\mathcal{E}$, the model refrained from annotating it; thus, demonstrating its cautious and appropriate decision-making in ambiguous cases.

Row (d) of  the figure shows a visually normal patch that received a preliminary \texttt{Unhealthy} label due to annotation noise. PEN embedded it near \texttt{Healthy} samples and SAFE confidently predicted it as \texttt{Healthy}, with the Grad-CAM-extracted visualization showing uniform activation across the patch. In Row (e), the input patch contained a small isolated microaneurysm. Despite its subtle appearance, the model was able to detect it successfully and embed the patch near another \texttt{Unhealthy} sample with a similar lesion. The Grad-CAM view shows that PEN focused precisely on the lesion; thereby, demonstrating the ability of the model to correctly identify subtle pathological features. Row (f) presents a patch with faint exudates, blurred vessels, and a placement of the lesion near the boundary. Although Grad-CAM results show high activation in the suspected lesion region, the model abstained from labeling it due to low confidence; thus reflecting a cautious attitude in uncertain scenario.

In general, the model demonstrated a strong semantic understanding of both \texttt{Healthy} and \texttt{Unhealthy} categories as well as finer-grained pathological characteristics, such as microaneurysms, exudates, etc. It consistently grouped the semantically similar patches in $\mathcal{E}$ and assigned labels with caution when evidence was weak. Its ability to abstain in ambiguous cases reinforces the reliability of the SAFE framework. This highlights its potential for trustworthy and interpretable weakly supervised annotation. 

\begin{table*}
\centering
\caption{Comparative analysis of annotation generated using SAFE in downstream task.}
\label{tab:res_downstream}
\resizebox{1.0\columnwidth}{!}{
\renewcommand{\arraystretch}{1.5}
\begin{tabular}{|l|l|c|c|c|c|c|ccc|ccc|}
\hline
\multirow{2}{*}{\textbf{Dataset}}          & \multirow{2}{*}{\textbf{Models}}          & \multirow{2}{*}{\textbf{SAFE}} & \multirow{2}{*}{\textbf{Acc $(\uparrow)$}} & \multirow{2}{*}{\textbf{BAcc $(\uparrow)$}} & \multirow{2}{*}{\textbf{AUPRC $(\uparrow)$}} & \multirow{2}{*}{\textbf{\begin{tabular}[c]{@{}c@{}}AUPRC\\ gain\end{tabular}}} & \multicolumn{3}{c|}{\textbf{Healthy}}                                                                                          & \multicolumn{3}{c|}{\textbf{Unhealthy}}                                                                                        \\ \cline{8-13} 
                                           &                                           &                                &                                            &                                             &                                              &                                                                                & \multicolumn{1}{c|}{\textbf{PR $(\uparrow)$}} & \multicolumn{1}{c|}{\textbf{RE $(\uparrow)$}} & \textbf{F1-score $(\uparrow)$} & \multicolumn{1}{c|}{\textbf{PR $(\uparrow)$}} & \multicolumn{1}{c|}{\textbf{RE $(\uparrow)$}} & \textbf{F1-score $(\uparrow)$} \\ \hline
\multirow{6}{*}{\textbf{Messidor$^{*}$}}   & \multirow{2}{*}{\textbf{ResNet18}}        & \bluecirc                      & 0.928 ± 0.002                              & 0.856 ± 0.003                               & 0.528 ± 0.008                                & \multirow{2}{*}{+0.159}                                                        & \multicolumn{1}{c|}{0.969 ± 0.000}            & \multicolumn{1}{c|}{0.948 ± 0.003}            & 0.959 ± 0.001                  & \multicolumn{1}{c|}{0.656 ± 0.012}            & \multicolumn{1}{c|}{0.764 ± 0.007}            & 0.706 ± 0.006                  \\ \cline{3-6} \cline{8-13} 
                                           &                                           & \bluecircgreendot              & \textbf{0.946 ± 0.000}                     & \textbf{0.903 ± 0.002}                      & \textbf{0.687 ± 0.002}                       &                                                                                & \multicolumn{1}{c|}{\textbf{0.974 ± 0.000}}   & \multicolumn{1}{c|}{\textbf{0.962 ± 0.001}}   & \textbf{0.968 ± 0.000}         & \multicolumn{1}{c|}{\textbf{0.788 ± 0.007}}   & \multicolumn{1}{c|}{\textbf{0.845 ± 0.006}}   & \textbf{0.815 ± 0.001}         \\ \cline{2-13} 
                                           & \multirow{2}{*}{\textbf{Inception-NetV3}} & \bluecirc                      & 0.911 ± 0.019                              & 0.867 ± 0.010                               & 0.498 ± 0.041                                & \multirow{2}{*}{+0.141}                                                        & \multicolumn{1}{c|}{0.974 ± 0.005}            & \multicolumn{1}{c|}{0.924 ± 0.027}            & 0.948 ± 0.012                  & \multicolumn{1}{c|}{0.592 ± 0.079}            & \multicolumn{1}{c|}{0.811 ± 0.046}            & 0.679 ± 0.035                  \\ \cline{3-6} \cline{8-13} 
                                           &                                           & \bluecircgreendot              & \textbf{0.932 ± 0.011}                     & \textbf{0.901 ± 0.004}                      & \textbf{0.639 ± 0.039}                       &                                                                                & \multicolumn{1}{c|}{\textbf{0.976 ± 0.003}}   & \multicolumn{1}{c|}{\textbf{0.944 ± 0.017}}   & \textbf{0.959 ± 0.007}         & \multicolumn{1}{c|}{\textbf{0.724 ± 0.065}}   & \multicolumn{1}{c|}{\textbf{0.857 ± 0.025}}   & \textbf{0.782 ± 0.025}         \\ \cline{2-13} 
                                           & \multirow{2}{*}{\textbf{ViT}}             & \bluecirc                      & 0.916 ± 0.001                              & 0.886 ± 0.000                               & 0.521 ± 0.001                                & \multirow{2}{*}{+0.138}                                                        & \multicolumn{1}{c|}{\textbf{0.979 ± 0.000}}   & \multicolumn{1}{c|}{0.926 ± 0.000}            & 0.952 ± 0.000                  & \multicolumn{1}{c|}{0.594 ± 0.001}            & \multicolumn{1}{c|}{0.847 ± 0.001}            & 0.699 ± 0.001                  \\ \cline{3-6} \cline{8-13} 
                                           &                                           & \bluecircgreendot              & \textbf{0.937 ± 0.001}                     & \textbf{0.913 ± 0.001}                      & \textbf{0.659 ± 0.004}                       &                                                                                & \multicolumn{1}{c|}{\textbf{0.979 ± 0.000}}   & \multicolumn{1}{c|}{\textbf{0.946 ± 0.001}}   & \textbf{0.962 ± 0.000}         & \multicolumn{1}{c|}{\textbf{0.729 ± 0.005}}   & \multicolumn{1}{c|}{\textbf{0.880 ± 0.001}}   & \textbf{0.798 ± 0.003}         \\ \hline
\multirow{6}{*}{\textbf{IDRiD$^{(-)}$}}    & \multirow{2}{*}{\textbf{ResNet18}}        & \bluecirc                      & 0.954 ± 0.001                              & 0.834 ± 0.011                               & 0.495 ± 0.011                                & \multirow{2}{*}{+0.107}                                                        & \multicolumn{1}{c|}{0.976 ± 0.001}            & \multicolumn{1}{c|}{0.974 ± 0.003}            & 0.975 ± 0.001                  & \multicolumn{1}{c|}{0.683 ± 0.025}            & \multicolumn{1}{c|}{0.694 ± 0.026}            & 0.688 ± 0.009                  \\ \cline{3-6} \cline{8-13} 
                                           &                                           & \bluecircgreendot              & \textbf{0.959 ± 0.001}                     & \textbf{0.872 ± 0.003}                      & \textbf{0.602 ± 0.010}                       &                                                                                & \multicolumn{1}{c|}{\textbf{0.978 ± 0.000}}   & \multicolumn{1}{c|}{\textbf{0.977 ± 0.001}}   & \textbf{0.977 ± 0.000}         & \multicolumn{1}{c|}{\textbf{0.759 ± 0.012}}   & \multicolumn{1}{c|}{\textbf{0.766 ± 0.008}}   & \textbf{0.763 ± 0.007}         \\ \cline{2-13} 
                                           & \multirow{2}{*}{\textbf{Inception-NetV3}} & \bluecirc                      & 0.931 ± 0.011                              & 0.864 ± 0.000                               & 0.425 ± 0.038                                & \multirow{2}{*}{+0.164}                                                        & \multicolumn{1}{c|}{\textbf{0.982 ± 0.000}}   & \multicolumn{1}{c|}{0.942 ± 0.013}            & 0.962 ± 0.006                  & \multicolumn{1}{c|}{0.521 ± 0.056}            & \multicolumn{1}{c|}{0.786 ± 0.013}            & 0.625 ± 0.037                  \\ \cline{3-6} \cline{8-13} 
                                           &                                           & \bluecircgreendot              & \textbf{0.956 ± 0.003}                     & \textbf{0.881 ± 0.006}                      & \textbf{0.589 ± 0.017}                       &                                                                                & \multicolumn{1}{c|}{0.980 ± 0.001}            & \multicolumn{1}{c|}{\textbf{0.971 ± 0.005}}   & \textbf{0.976 ± 0.002}         & \multicolumn{1}{c|}{\textbf{0.723 ± 0.037}}   & \multicolumn{1}{c|}{\textbf{0.790 ± 0.019}}   & \textbf{0.755 ± 0.011}         \\ \cline{2-13} 
                                           & \multirow{2}{*}{\textbf{ViT}}             & \bluecirc                      & 0.929 ± 0.023                              & 0.854 ± 0.025                               & 0.429 ± 0.085                                & \multirow{2}{*}{+0.134}                                                        & \multicolumn{1}{c|}{\textbf{0.981 ± 0.005}}   & \multicolumn{1}{c|}{0.942 ± 0.031}            & 0.961 ± 0.013                  & \multicolumn{1}{c|}{0.553 ± 0.181}            & \multicolumn{1}{c|}{\textbf{0.766 ± 0.081}}   & 0.622 ± 0.068                  \\ \cline{3-6} \cline{8-13} 
                                           &                                           & \bluecircgreendot              & \textbf{0.953 ± 0.012}                     & \textbf{0.854 ± 0.022}                      & \textbf{0.563 ± 0.051}                       &                                                                                & \multicolumn{1}{c|}{0.975 ± 0.005}            & \multicolumn{1}{c|}{\textbf{0.973 ± 0.019}}   & \textbf{0.974 ± 0.007}         & \multicolumn{1}{c|}{\textbf{0.742 ± 0.115}}   & \multicolumn{1}{c|}{0.735 ± 0.063}            & \textbf{0.730 ± 0.034}         \\ \hline
\multirow{6}{*}{\textbf{e-ophtha$^{(-)}$}} & \multirow{2}{*}{\textbf{ResNet18}}        & \bluecirc                      & 0.946 ± 0.003                              & 0.785 ± 0.022                               & 0.194 ± 0.018                                & \multirow{2}{*}{+0.204}                                                        & \multicolumn{1}{c|}{0.988 ± 0.001}            & \multicolumn{1}{c|}{0.956 ± 0.004}            & 0.972 ± 0.002                  & \multicolumn{1}{c|}{0.298 ± 0.019}            & \multicolumn{1}{c|}{0.613 ± 0.046}            & 0.401 ± 0.021                  \\ \cline{3-6} \cline{8-13} 
                                           &                                           & \bluecircgreendot              & \textbf{0.970 ± 0.009}                     & \textbf{0.861 ± 0.008}                      & \textbf{0.398 ± 0.053}                       &                                                                                & \multicolumn{1}{c|}{\textbf{0.990 ± 0.000}}   & \multicolumn{1}{c|}{\textbf{0.973 ± 0.007}}   & \textbf{0.981 ± 0.003}         & \multicolumn{1}{c|}{\textbf{0.523 ± 0.071}}   & \multicolumn{1}{c|}{\textbf{0.750 ± 0.018}}   & \textbf{0.613 ± 0.045}         \\ \cline{2-13} 
                                           & \multirow{2}{*}{\textbf{Inception-NetV3}} & \bluecirc                      & 0.955 ± 0.006                              & 0.792 ± 0.016                               & 0.230 ± 0.026                                & \multirow{2}{*}{+0.191}                                                        & \multicolumn{1}{c|}{0.988 ± 0.001}            & \multicolumn{1}{c|}{0.965 ± 0.007}            & 0.977 ± 0.003                  & \multicolumn{1}{c|}{0.355 ± 0.042}            & \multicolumn{1}{c|}{0.618 ± 0.035}            & 0.449 ± 0.034                  \\ \cline{3-6} \cline{8-13} 
                                           &                                           & \bluecircgreendot              & \textbf{0.970 ± 0.001}                     & \textbf{0.847 ± 0.012}                      & \textbf{0.421 ± 0.012}                       &                                                                                & \multicolumn{1}{c|}{\textbf{0.989 ± 0.001}}   & \multicolumn{1}{c|}{\textbf{0.979 ± 0.002}}   & \textbf{0.984 ± 0.000}         & \multicolumn{1}{c|}{\textbf{0.574 ± 0.016}}   & \multicolumn{1}{c|}{\textbf{0.715 ± 0.027}}   & \textbf{0.637 ± 0.009}         \\ \cline{2-13} 
                                           & \multirow{2}{*}{\textbf{ViT}}             & \bluecirc                      & 0.882 ± 0.027                              & 0.815 ± 0.017                               & 0.133 ± 0.024                                & \multirow{2}{*}{+0.316}                                                        & \multicolumn{1}{c|}{\textbf{0.991 ± 0.000}}   & \multicolumn{1}{c|}{0.886 ± 0.027}            & 0.936 ± 0.015                  & \multicolumn{1}{c|}{0.168 ± 0.032}            & \multicolumn{1}{c|}{\textbf{0.743 ± 0.019}}   & 0.274 ± 0.043                  \\ \cline{3-6} \cline{8-13} 
                                           &                                           & \bluecircgreendot              & \textbf{0.974 ± 0.003}                     & \textbf{0.830 ± 0.007}                      & \textbf{0.449 ± 0.043}                       &                                                                                & \multicolumn{1}{c|}{0.987 ± 0.000}            & \multicolumn{1}{c|}{\textbf{0.985 ± 0.004}}   & \textbf{0.986 ± 0.002}         & \multicolumn{1}{c|}{\textbf{0.648 ± 0.062}}   & \multicolumn{1}{c|}{0.675 ± 0.014}            & \textbf{0.660 ± 0.033}         \\ \hline
\multirow{6}{*}{\textbf{DDR$^{(-)}$}}      & \multirow{2}{*}{\textbf{ResNet18}}        & \bluecirc                      & 0.926 ± 0.003                              & 0.858 ± 0.000                               & 0.316 ± 0.009                                & \multirow{2}{*}{\textbf{+0.538}}                                               & \multicolumn{1}{c|}{0.987 ± 0.000}            & \multicolumn{1}{c|}{0.933 ± 0.004}            & 0.959 ± 0.002                  & \multicolumn{1}{c|}{0.390 ± 0.014}            & \multicolumn{1}{c|}{0.782 ± 0.004}            & 0.520 ± 0.011                  \\ \cline{3-6} \cline{8-13} 
                                           &                                           & \bluecircgreendot              & \textbf{0.971 ± 0.001}                     & \textbf{0.960 ± 0.000}                      & \textbf{0.854 ± 0.008}                       &                                                                                & \multicolumn{1}{c|}{\textbf{0.988 ± 0.000}}   & \multicolumn{1}{c|}{\textbf{0.976 ± 0.002}}   & \textbf{0.982 ± 0.001}         & \multicolumn{1}{c|}{\textbf{0.894 ± 0.010}}   & \multicolumn{1}{c|}{\textbf{0.944 ± 0.001}}   & \textbf{0.918 ± 0.004}         \\ \cline{2-13} 
                                           & \multirow{2}{*}{\textbf{Inception-NetV3}} & \bluecirc                      & 0.951 ± 0.052                              & 0.894 ± 0.049                               & 0.301 ± 0.014                                & \multirow{2}{*}{\textbf{+0.545}}                                               & \multicolumn{1}{c|}{0.988 ± 0.000}            & \multicolumn{1}{c|}{0.925 ± 0.006}            & 0.955 ± 0.003                  & \multicolumn{1}{c|}{0.365 ± 0.018}            & \multicolumn{1}{c|}{0.796 ± 0.001}            & 0.500 ± 0.016                  \\ \cline{3-6} \cline{8-13} 
                                           &                                           & \bluecircgreendot              & \textbf{0.969 ± 0.002}                     & \textbf{0.960 ± 0.001}                      & \textbf{0.846 ± 0.010}                       &                                                                                & \multicolumn{1}{c|}{\textbf{0.989 ± 0.001}}   & \multicolumn{1}{c|}{\textbf{0.974 ± 0.004}}   & \textbf{0.981 ± 0.001}         & \multicolumn{1}{c|}{\textbf{0.884 ± 0.016}}   & \multicolumn{1}{c|}{\textbf{0.947 ± 0.007}}   & \textbf{0.914 ± 0.005}         \\ \cline{2-13} 
                                           & \multirow{2}{*}{\textbf{ViT}}             & \bluecirc                      & 0.964 ± 0.001                              & 0.797 ± 0.001                               & 0.424 ± 0.010                                & \multirow{2}{*}{+0.433}                                                        & \multicolumn{1}{c|}{0.979 ± 0.000}            & \multicolumn{1}{c|}{0.983 ± 0.001}            & 0.981 ± 0.000                  & \multicolumn{1}{c|}{0.661 ± 0.015}            & \multicolumn{1}{c|}{0.612 ± 0.000}            & 0.635 ± 0.007                  \\ \cline{3-6} \cline{8-13} 
                                           &                                           & \bluecircgreendot              & \textbf{0.971 ± 0.001}                     & \textbf{0.964 ± 0.022}                      & \textbf{0.857 ± 0.007}                       &                                                                                & \multicolumn{1}{c|}{\textbf{0.982 ± 0.001}}   & \multicolumn{1}{c|}{\textbf{0.984 ± 0.002}}   & \textbf{0.983 ± 0.000}         & \multicolumn{1}{c|}{\textbf{0.922 ± 0.011}}   & \multicolumn{1}{c|}{\textbf{0.913 ± 0.005}}   & \textbf{0.918 ± 0.003}         \\ \hline
\end{tabular}}
\end{table*}

\subsection{Impact of SAFE annotation in downstream task}
\label{downstream}

This section evaluates the annotations generated by SAFE in downstream fundus patch classification, as reported in Table~\ref{tab:res_downstream}. In the ``without-SAFE" (\bluecirc) protocol, all \texttt{Unlabeled} patches get absorbed into the \texttt{Healthy} category. This reflects the default behavior of the DL models. In the ``with-SAFE" (\bluecircgreendot) protocol, SAFE assigns \texttt{Unlabeled} patches as \texttt{Healthy} or \texttt{Unhealthy}, while excluding those marked by it as \texttt{Undecided}. The ``without-" and ``with-SAFE" protocols apply only to the training data. 
Deep models ResNet18 \citep{resnet}, Inception-NetV3 \citep{inception}, and ViT \citep{vit} were independently trained under both protocols, using the train data, and evaluated on the previous holdout \texttt{Labeled} test split. This design isolates the impact of SAFE annotations for reflecting their effectiveness and reliability.

\noindent \textit{{\bf Messidor$^{*}$}}: The SAFE protocol produced consistent gains across all models, with the highest improvement in \texttt{Unhealthy} class metrics and AUPRC. Precision and recall in the \texttt{Healthy} class increased slightly, with a modest improvement in F1-score. Substantial gains in the \texttt{Unhealthy} class indicate that SAFE correctly segregated additional \texttt{Unhealthy} patches from the \texttt{Unlabeled} set, while excluding ambiguous samples.

\noindent \textit{{\bf IDRiD$^{(-)}$}}: It matches Messidor$^{*}$ in dataset size and class imbalance, but comes with more detailed lesion-level annotations; thereby, making the baseline (without SAFE) annotations stronger. Consequently, the SAFE annotations yielded moderate improvement. The SAFE protocol delivered higher overall F1-scores despite without-SAFE showing marginally higher \texttt{Healthy} precision and recall with Inception-NetV3 and ViT. The consistent increase in AUPRC implies that SAFE improved the detection of \texttt{Unhealthy} patches, even in the presence of strong baseline labels.

\noindent \textit{{\bf e-ophtha$^{(-)}$}}: The SAFE protocol delivered strong improvement despite the small size of the dataset. The \texttt{Healthy} class already performed well without the SAFE protocol, and showed comparable or slightly higher performance under the SAFE protocol. The \texttt{Unhealthy} class improved substantially across all models, with clear gains in precision, recall, and F1-score. As claimed in Section~\ref{subsec:ComAna}, SAFE proves most critical here by enriching the sparse \texttt{Unhealthy} representation, and reducing noise through the exclusion of uncertain patches.

\noindent \textit{{\bf DDR$^{(-)}$}}: This showed the most pronounced and stable improvement over all metrics. The strong scores, observed in Table \ref{tab:ablation} in DDR$^{(-)}$ for the combined loss $\mathcal{L}$, likely contributed to the effective annotation produced by SAFE. Both classes achieved significantly increased balanced performance, resulting in high balanced accuracy. The large increase in AUPRC indicates that SAFE contributes to a substantial number of correct \texttt{Unhealthy} labels rather than noisy ones. These improvements confirm that SAFE scales effectively to large datasets, and addresses class imbalance while maintaining label quality.

In summary, SAFE consistently improved Balanced Accuracy and AUPRC for the \texttt{Healthy} category across all datasets and models; while for the \texttt{Unhealthy} category the improvement was observed over all the metrics involved. As the \texttt{Healthy} class already performed strongly without SAFE protocol, therefore the gains remained smaller but positive; thus confirming that the SAFE annotations preserved semantic representation. The \texttt{Unhealthy} class demonstrated the highest improvement, with substantial gains in both precision and recall. Note that a higher recall reflects meaningful addition of \texttt{Unhealthy} samples, while stable or improved precision indicates that SAFE does not introduce significant noise. This confirms that SAFE fulfills its intended role of efficiently annotating previously unlabeled patches.

\subsection{Sensitivity analysis}
\label{subsec:senanal}

This section evaluates the robustness of the SAFE framework and the selection of its hyperparameters through an extensive sensitivity analysis conducted on the Messidor$^{*}$ dataset, across the parameters $\lambda$, $K$ and $\tau$.

\begin{figure*}[ht]
    \centering
    \begin{subfigure}{0.49\textwidth}
        \centering
        \includegraphics[width=\linewidth]{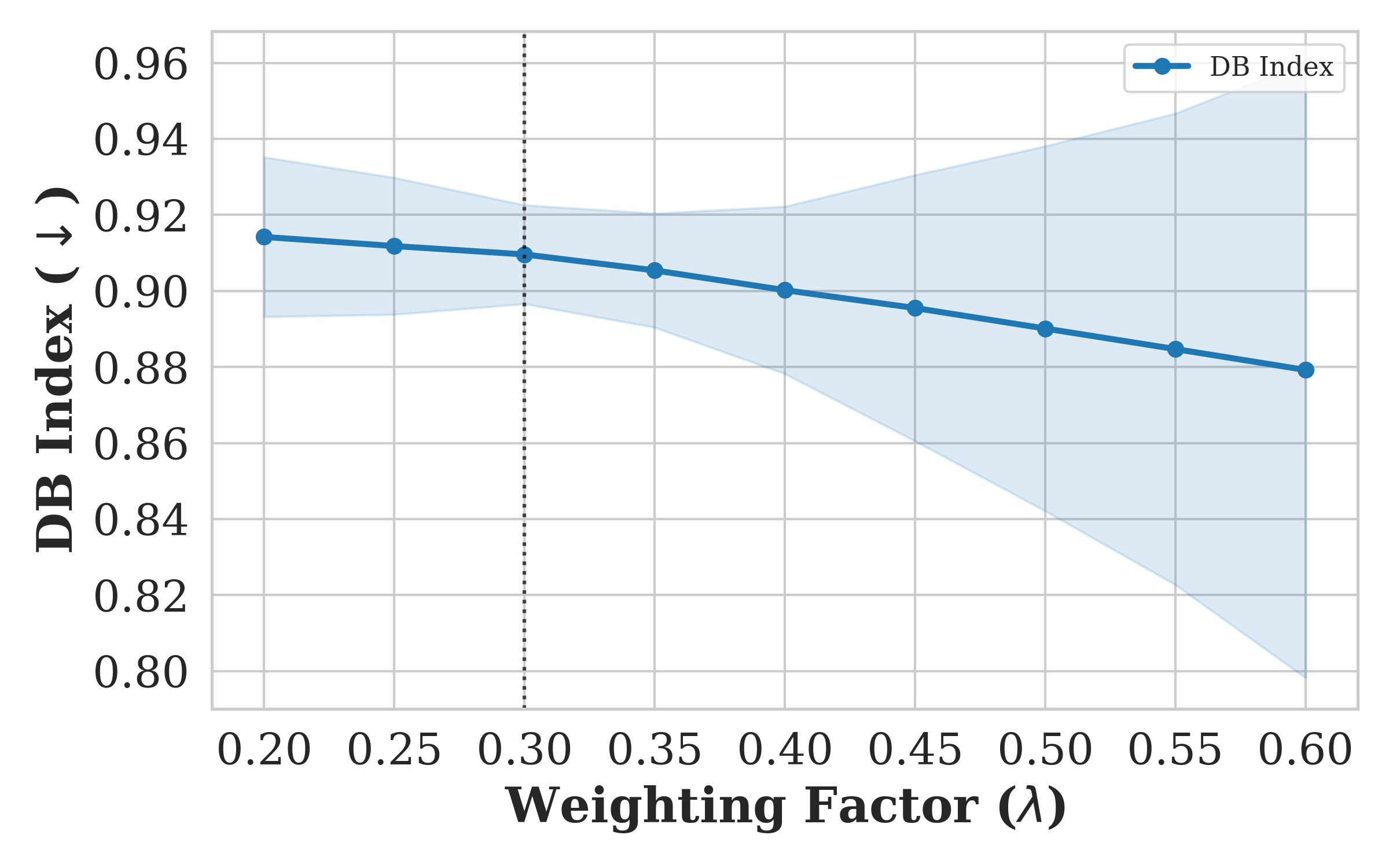}
        \caption{Clustering Compactness (DB Index)}
        \label{fig:senana_lambda_db}
    \end{subfigure}
    \hfill
    \begin{subfigure}{0.49\textwidth}
        \centering
        \includegraphics[width=\linewidth]{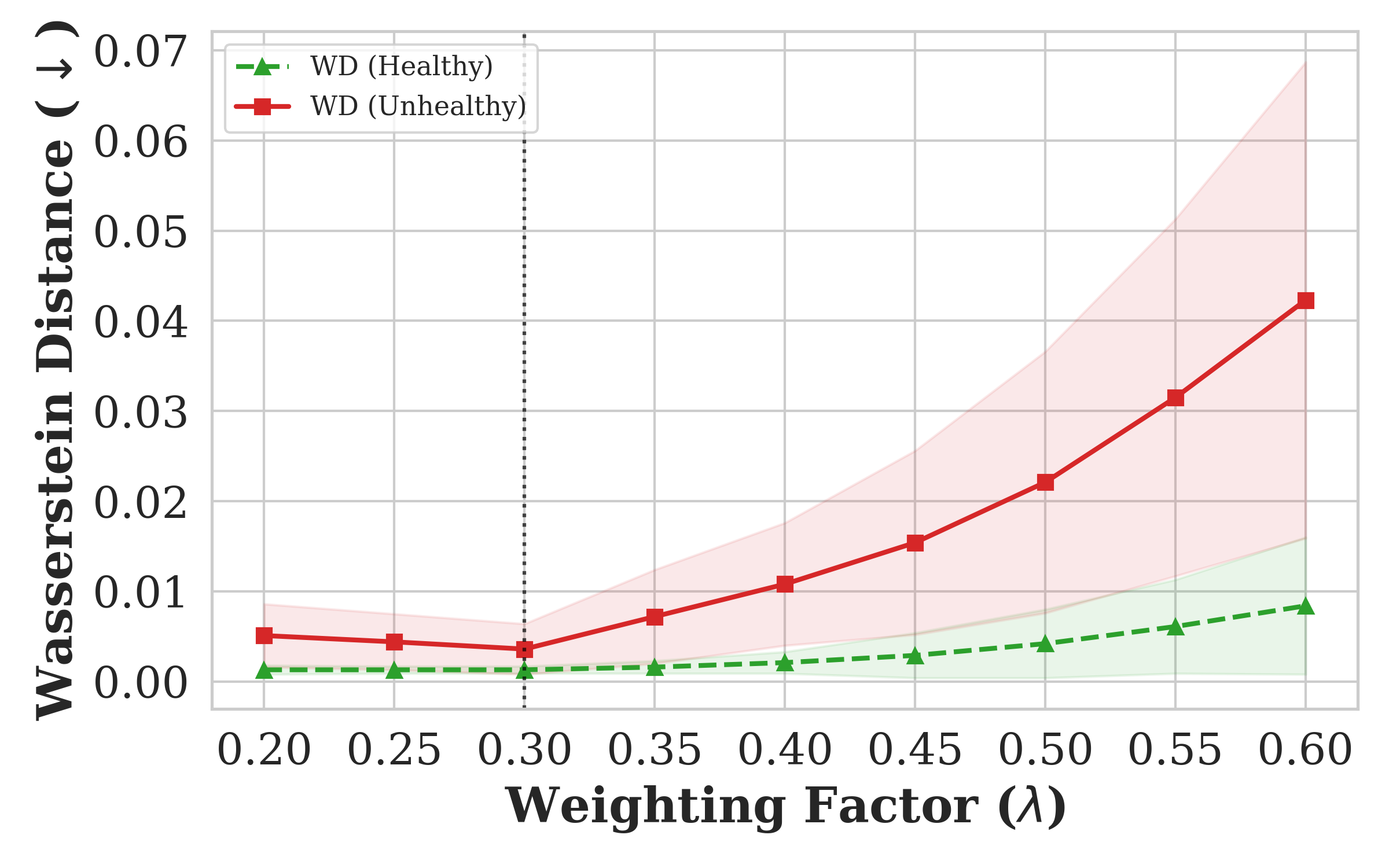}
        \caption{Distributional Alignment (WD)}
        \label{fig:senana_lambda_wd}
    \end{subfigure}
    \caption{Sensitivity analysis of the weighting factor $\lambda$ on the Messidor$^{*}$ dataset.}
    \label{fig:senana_lambda}
\end{figure*}

\paragraph{Analysis on $\lambda$} Fig. \ref{fig:senana_lambda} reveals a critical trade-off between the compactness and the distributional robustness of the feature embedding spaces for the Messidor$^{*}$ data. As $\lambda$ increases from 0.2 to 0.6, the DB-index gets consistently decreased. This indicates that a strong $\mathcal{L}_\text{BCE}$ produces tighter clusters. However, the gain triggers a simultaneous increase in WD, particularly for the \texttt{Unhealthy} class. This suggests that over-reliance on $\mathcal{L}_\text{BCE}$ compromises alignment with the unseen distribution. Thus, setting $\lambda=0.3$ identifies the optimal equilibrium by achieving a low DB index while maintaining the minimum WD score. For smaller or noisier datasets, like e-ophtha or DDR, $\lambda=0.3$ demonstrated better cluster formation, with  $\mathcal{L}_\text{SCL}$ enforcing semantic understanding while reducing  complete dependence on the unreliable preliminary labels.

\paragraph{Analysis on $K$ and $\tau$}
\begin{figure}[ht]
    \centering
    \includegraphics[width=0.6\columnwidth]{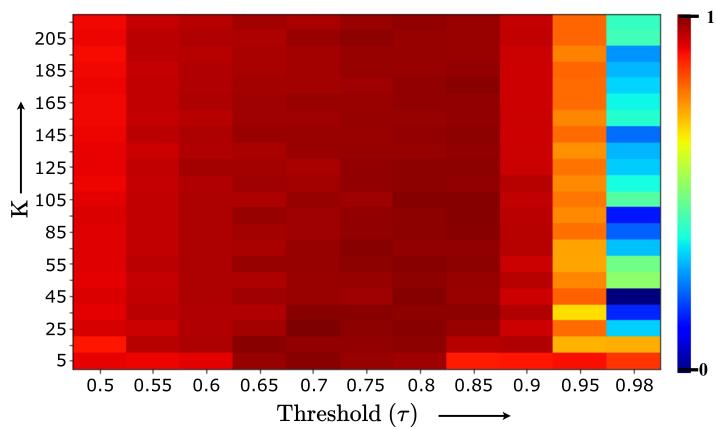}
    \caption{Sensitivity analysis across $K$--$\tau$ configurations on the Messidor$^{*}$ dataset.}    \label{fig:senana_messidor}
\end{figure}

The Polygon Area Metric (PAM)~\citep{aydemir2021new}, used to analyze sensitivity across $K$--$\tau$ configurations, jointly summarizes multiple evaluation metrics into a single scalar value. Fig.~\ref{fig:senana_messidor} shows the normalized PAM heatmap for the Messidor dataset, highlighting the configuration that achieved the highest overall performance. It improved with moderate $K$, which provided reliable neighborhood information. However, excessively large $K$ values diluted the local structure and slightly reduced the annotation quality. The parameter $\tau$ had the strongest influence, with very high values ($>0.85$) degrading the performance by sharply reducing $D_\text{rate}$ despite improving annotation quality. Very low values ($<0.65$) increased $D_\text{rate}$ but lowered scores in the rest of the evaluation metric. These observations confirm the benefit of maintaining a flexible $\tau$ within the SAFE framework. The unified scalar projection (by PAM) is also used to identify the most appropriate $K$--$\tau$ configuration, such that error is minimized in each dataset for the framework.

\subsection{Qualitative validation of SAFE-generated annotations}
\label{subsec:annotation_validation}

\begin{figure}[ht]
    \centering
    \includegraphics[width=0.8\columnwidth]{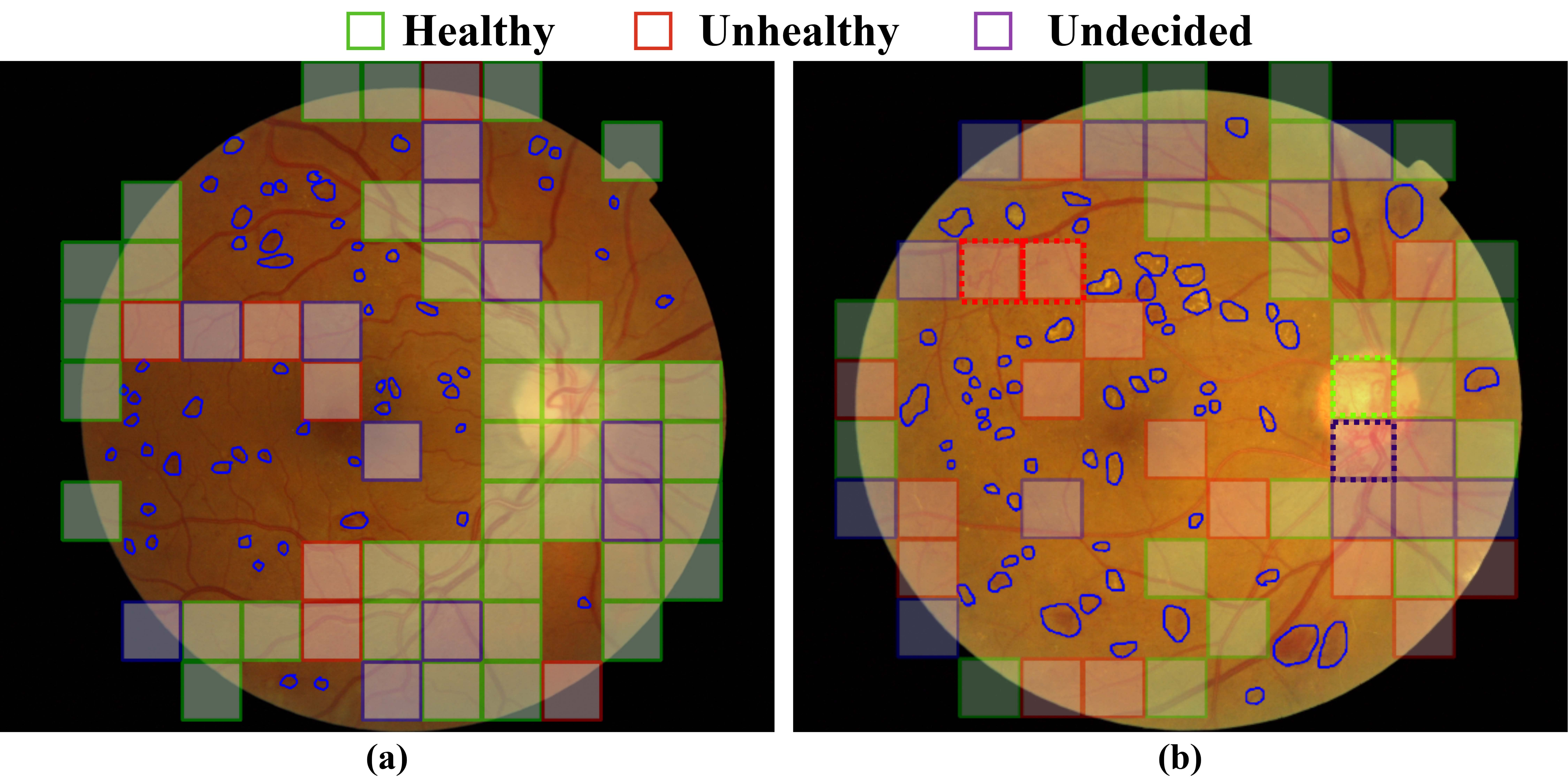}
    \caption{Representative fundus images, annotated with SAFE-generated patch labels, highlighting (a) the strength, and (b) some limitations of the algorithm.}
    \label{fig:annotation_full}
\end{figure}

Fig.~\ref{fig:annotation_full} presents two sample fundus images, with SAFE-generated annotations overlayed across all the patches. In  part (a) of the figure, SAFE is found to correctly identify all unlabeled patches as either \texttt{Healthy} or \texttt{Unhealthy}, while appropriately labeling ambiguous low-contrast boundary regions as \texttt{Undecided}. Part (b) of the figure shows that SAFE correctly identified intraretinal microvascular abnormalities (IRMA) as \texttt{Unhealthy} (as shown by the red dotted box). Neovascularization proved more challenging, with one neovascularization of the disc (NVD) being mislabeled as \texttt{Healthy}  (depicted by the green dotted box) and neovascularization elsewhere (NVE) being conservatively marked as \texttt{Undecided}  (as indicated by the violet dotted box). Neovascularization is typically underrepresented in datasets and appears as small, irregular, and faint vascular structures having low contrast. The bright optic disc background further reduced vessel–background contrast for NVD to make detection harder than in NVE,  over the darker retinal regions.

SAFE accurately highlighted regions affected by lesions, such as hemorrhages, exudates, and IRMA, while maintaining robustness through the \texttt{Undecided} category in ambiguous patches. Misannotations were limited, occurring mainly in peripheral regions or in proliferative DR lesions such as NVD (which are inherently difficult to detect). The cautious labeling of NVE as \texttt{Undecided} reflected recognition of abnormality, and the occasional NVD miss corresponded to the inherent difficulty of detecting such low-contrast lesions. 

The clinical experts in the team validated the annotated images, confirming that the SAFE-generated annotations conformed to medical judgment; thereby, reinforcing its practical utility.

\section{Conclusion}
\label{sec:conclusion}

The article presented the design and development of SAFE, a two-stage novel framework for generating reliable patch-level annotations from weakly supervised fundus images to support automated DR screening. The method used ensemble of contrastive embedding spaces, with a proximity based approach, to infer labels for the \texttt{Unlabeled} patches. This helped in addressing the challenge of incomplete lesion annotations prevalent in DR datasets. SAFE-generated annotations achieved high alignment with ground-truth labels. Quantitative evaluation demonstrated that SAFE outperformed the baseline by producing annotations that significantly improved downstream classification; particularly, for the under-represented \texttt{Unhealthy} class. The gains were most pronounced in the highly imbalanced datasets, such as Messidor, e-ophtha and DDR, having varying sizes. Moderate improvements were observed in IDRiD, which had comprehensive lesion labeling.

The design of the SAFE framework facilitates its extension to other medical domains, such as histopathology. SAFE is particularly suited for use in high-resolution, weakly labeled datasets, where localized abnormalities need to be inferred with minimal supervision. Although Stage 2 of the SAFE framework provides high precision annotation, the integration of efficient similarity search architectures (such as hierarchical navigable small world or approximate nearest neighbor framework like FAISS) can further improve the scalability. Future research can also explore active learning strategies to identify and prioritize uncertain patches for expert validation; thereby, minimizing annotation effort while improving quality. 

\section*{Acknowledgement}

Abhirup Banerjee is supported by the Royal Society University Research Fellowship (Grant No. URF/R1/221314).  
Research of Sushmita Mitra is funded by the J. C. Bose National Fellowship (Grant No. JCB/2020/000033).

\bibliographystyle{unsrt}  
\bibliography{references}

\end{document}